\pdfoutput=1
\documentclass[11pt]{article}

\usepackage{acl}

\usepackage{times}
\usepackage{latexsym}

\usepackage[T1]{fontenc}
\usepackage[utf8]{inputenc}

\usepackage{microtype}
\usepackage{booktabs}       % professional-quality tables
\usepackage{amsfonts}       % blackboard math symbols
\usepackage{nicefrac}       % compact symbols for 1/2, etc.
\usepackage{microtype}      % microtypography
\usepackage{xcolor}         % colors

\usepackage{epsfig}
\usepackage{graphicx}
\usepackage{amsmath}
\usepackage{amssymb}
\usepackage{xspace}
\usepackage{amstext}
\usepackage{multirow}
\usepackage{booktabs}   
\usepackage{subcaption}
\usepackage{colortbl}
\usepackage{hyperref}
\usepackage{todonotes}
\usepackage{placeins}
\usepackage{wrapfig}
\usepackage{longtable}
\newcommand{\bin}{Babel-ImageNet}
\newcommand{\repourl}{https://github.com/gregor-ge/Babel-ImageNet} %\url{https://github.com/gregor-ge/Babel-ImageNet}} 

\title{Babel-ImageNet: Massively Multilingual Evaluation of Vision-and-Language Representations}

\author{\textbf{Gregor Geigle$^{12}$ \quad \quad Radu Timofte$^{2}$ \quad \quad Goran Glava\v{s}$^{1}$} \\
  $^{1}$W{\"u}NLP, $^{2}$Computer Vision Lab, CAIDAS, University of W{\"u}rzburg,  \\
  \texttt{gregor.geigle@uni-wuerburg.de}}

\begin{document}
\maketitle

\begin{abstract}
% CLIP-style multimodal have proven useful in bridging the modality gap between text and images.
% However, evaluation is mainly in English with ImageNet zero-shot as the prime metric.
% Multilingual evaluation is limited to image-text retrieval, but this requires human annotators, making this costly to scale to many languages.
% As a results, current coverage is mostly for only a few high-resource languages.

% We present Babel-ImageNet, which provides partial translations of the 1000 ImageNet labels to 92 languages without human annotators or machine translation.
% Instead, we leverage the structural connection between ImageNet, WordNet synsets and BabelNet, a multilingual semantic net, to attain clean translations automatically.

% We evaluate 8 publicly available multilingual CLIP models with our benchmark and show that there is a large gap between English ImageNet results and even high-resource languages like German or Chinese and an even larger gap to low-resource languages.
% We extensively validate our benchmark for evaluation of multilingual multimodal representations and show that our results highly correlate with retrieval results (while being far easier to procure for many languages) and that even with only partial translations, our results are a meaningful estimate of performance.

Vision-and-language (VL) models with separate encoders for each modality (e.g., CLIP) have become the go-to models for zero-shot image classification and image-text retrieval. 
They are, however, mostly evaluated in English as multilingual benchmarks are limited in availability.
We introduce Babel-ImageNet, a massively multilingual benchmark that offers (partial) translations of ImageNet labels to 100 languages, built without machine translation or manual annotation. We instead automatically obtain reliable translations by linking them -- via shared WordNet synsets -- to BabelNet, a massively multilingual lexico-semantic network.
We evaluate 11 public multilingual CLIP models on zero-shot image classification (ZS-IC) on our benchmark, demonstrating a significant gap between English ImageNet performance and that of high-resource languages (e.g., German or Chinese), and an even bigger gap for low-resource languages (e.g., Sinhala or Lao). 
Crucially, we show that the models' ZS-IC performance 
highly correlates with their performance in image-text retrieval, validating the use of \bin{} to evaluate multilingual models for the vast majority of languages without gold image-text data.
Finally, we show that the performance of multilingual CLIP can be drastically improved for low-resource languages with parameter-efficient language-specific training. 
We make our code and data publicly available: \url{\repourl}
\end{abstract}

\section{Introduction}
% CLIP-like models that have been trained on web-scale image-caption data to encode the two modalities in a joint embedding space learn widely useful representations for both modalities \citep{radford_learning_2021,jia_scaling_2021,pham_combined_2021}.
% Those representations can be used directly for tasks like image-text retrieval \citep{lin_microsoft_2014,plummer_flickr30k_2015} or zero-shot image classification \citep{radford_learning_2021} but they can also be used as input encoding for downstream tasks like image generation \citep{rombach_high-resolution_2022} or joint cross-modal reasoning \citep{eichenberg_magma_2021,li_blip-2_2023}.

CLIP models \citep{radford_learning_2021,jia_scaling_2021,pham_combined_2021} have become widely used vision-and-language (VL) models, owing popularity to 
efficient inference based on separate yet semantically aligned encoders for the two modalities. Their bi-encoder architecture makes them ideal for efficient 
%
%Models that use CLIP-like methods to train on web-scale image-caption data and encode the two modalities in a joint embedding space learn versatile representations for both modalities~. These representations can be applied directly to tasks such as 
image-text retrieval~\citep{lin_microsoft_2014,plummer_flickr30k_2015} and zero-shot image classification~\citep{radford_learning_2021}, or to produce downstream features for supervised tasks such as image generation~\citep{rombach_high-resolution_2022} or cross-modal reasoning~\citep{eichenberg_magma_2021,li_blip-2_2023}.

% In general, CLIP models are evaluated on a diverse suit of image classification datasets in the zero-shot setting with ImageNet \citep{deng_imagenet_2009} as the \textit{main} dataset, motivated by findings that show good image representations for ImageNet transfer well to other general image problems (with specialized domains like medical imagery excluded) \citep{recht_imagenet_2019,fang_does_2023}.
% Implicitly, as class labels are generally only available in English, this evaluation method is limited to CLIP models trained with English caption data only.

Motivated by the observation that performance on ImageNet classification translates well to performance in many other image tasks~\citep{recht_imagenet_2019,fang_does_2023}, 
%(i.e., excluding specialized domains like medical imagery)
CLIP models are typically evaluated on zero-shot image classification (ZS-IC), i.e., by comparing the representation of an image with text representations of class labels, whereby ImageNet~\citep{deng_imagenet_2009} is the most prominent benchmark.
With ImageNet class labels available only in English, this supports only evaluation of monolingual English models (i.e., models trained with English captions only). 
%%
% Though most CLIP models are trained on English data, there have been efforts to train multilingual and monolingual non-English models both from scratch or via teacher distillation from English models \citep{carlsson_cross-lingual_2022,ilharco_openclip_2021,chen_altclip_2022,yang_chinese_2022,bianchi_contrastive_2021,zhang_generalizing_2022}, which opens up the aforementioned applications of CLIP representations to non-English speakers.
% However, while efforts have been made to translate ImageNet labels (and labels from other datasets) to a handfull of other languages \citep{bianchi_contrastive_2021,yang_chinese_2022}, language coverage for class labels is still very limited.
% Instead, multilingual CLIP models have mainly been evaluated using image-text retrieval datasets (e.g. \citep{aggarwal_towards_2020}), which have a limited coverage of mostly mid-to-high resource languages.
%%
Although most CLIP models are trained on English-only image-caption data, some effort has been put into creating multilingual and monolingual non-English models by (1) training them from scratch \citep{bianchi_contrastive_2021,ilharco_openclip_2021,yang_chinese_2022,jain_mural_2021,zhai_sigmoid_2023} or (2) distilling them from English models \citep{carlsson_cross-lingual_2022,chen_altclip_2022,zhang_generalizing_2022,visheratin_nllb_2023}, typically using parallel data as supervision. %, supporting VL representations for languages other than English. 
Despite attempts to translate ImageNet labels 
%(and labels from other datasets) 
to other languages~\citep{bianchi_contrastive_2021,yang_chinese_2022}, the language coverage 
%for ImageNet class labels 
remains very limited. Because of this, multilingual CLIP models have mainly been benchmarked on image-text retrieval datasets \citep{aggarwal_towards_2020,bugliarello_iglue_2022,thapliyal_crossmodal-3600_2022}, which cover only limited sets of mid-to-high resource languages.

%%% Cut for now since we need space
% \begin{figure}
%     \centering
%     \includegraphics[width=0.4\textwidth]{figures/wordcloud.pdf}
%     \caption{Babel-ImageNet translations in 92 languages for the ImageNet class 309 (synset n02206856: ``bee''). Font size of the language is proportional to the number of translated ImageNet classes. [Image by Charles J. Sharp, \href{https://creativecommons.org/licenses/by-sa/3.0}{CC BY-SA 3.0}, via \href{https://commons.wikimedia.org/wiki/File:Honey_bee_(Apis_mellifera).jpg}{Wikimedia Commons}]
%     }
%     %\vspace{-1mm}
% \label{fig:dataset_illustration}
% %\vspace{-3mm}
% \end{figure}

% \noindent \textbf{Rationale.} 
%%%%
Creating \textit{massively multilingual} %gold-standard 
datasets for VL tasks (e.g., image-text retrieval) 
%for very many languages (i.e., \textit{massively multilingual})
is prohibitively expensive. Existing efforts \citep{aggarwal_towards_2020,bugliarello_iglue_2022,thapliyal_crossmodal-3600_2022} either hire native speakers to write image captions in target languages or resort to machine translation (MT) of English data, followed by manual post-editing by native speakers. 
%%%
The MT approach (the cheaper of the two), is, we argue, still too expensive for low-resource languages because MT models are less accurate when translating to those languages, which implies a bigger post-editing effort for bilingual annotators, native in the low-resource language and fluent in English; in addition, such annotators are more difficult to find for low-resource than for high-resource target languages. 
% (compare, e.g., Sinhala and German). 
%\gregor{Need to also address the "write captions from scratch" approach. Findings speakers is likely still the main problem, I suppose?} GG: well no, that's even more expensive. Translation+post-editing is the cheaper variant, but still too expensive for low-resource languages. 
%%
In this work, we thus seek to create a robust \textit{massively multilingual benchmark} for evaluating the quality of representation spaces of multilingual VL models, without resorting to MT or requiring any manual annotation effort. To be useful, such a benchmark needs to satisfy a crucial requirement: models' performance across languages must be indicative of their performance for the same languages in tasks such as image-text retrieval, for which creating massively multilingual (gold-standard) evaluation datasets is too expensive.     

\noindent \textbf{Contributions.} With this in mind, we create \bin, a massively multilingual dataset for zero-shot image classification that offers (partial) translations of the 1000 ImageNet classes to 100 languages. To obtain robust translations of ImageNet labels in other languages, we leverage the connection between ImageNet classes, which are derived from WordNet~\citep{miller_wordnet_1994} synsets, and BabelNet~\citep{navigli_babelnet_2010}, a massively multilingual lexico-semantic network, also (in part) derived from WordNet.  
%%Given that iTo this end, we create a massively multilingual benchmark for zero-shot image classification  choose zero-shot image classification a To bridge this gap and enable zero-shot evaluation of multilingual CLIP models with ImageNet, we create Babel-ImageNet, a dataset that offers partial translations of the 1000 ImageNet-1k classes to 92 languages.
%%
Relying on the multilingual BabelNet synsets (and WordNet synset identifiers of ImageNet classes) to pivot between languages, we avoid problems known to occur with machine translation of short phrases without context, e.g., due to polysemy\footnote{For example, the ImageNet class \textit{walking stick} refers to the \textit{insect} and not the \textit{inanimate object}.}. Exploiting BabelNet allows us to automatically obtain labels for ImageNet concepts in many languages, removing the need for MT and manual annotation.

%Our contributions are as follows:
%1) We introduce Babel-ImageNet, a dataset that provides partial translations of the 1000 ImageNet classes through BabelNet to 92 languages, which allows us to test multilingual image-text models on ImageNet with the widest range of languages so far.
%%
We evaluate 11 different multilingual CLIP models on \bin, observing that all of them exhibit poor performance for low-resource languages. 
Crucially, we validate that \bin{} is a meaningful benchmark for measuring the quality of multilingual VL representations by comparing models' performance on \bin{} against their performance on established multilingual image-text retrieval datasets.   %% 
\bin{} thus allows us to evaluate models in languages not covered by those datasets and it additionally expands the retrieval-focused evaluation with the ZS-IC task in languages included in the established datasets.
Finally, we propose a computationally efficient approach for improving multilingual CLIP models for low-resource languages. This modular language specialization approach yields large performance gains ($>$20\% for some of the low-resource languages). 
%, investigating different training objectives. This language-specialization training these strategies are especially  For low-resource languages, we see huge improvements (>20\% accuracy) but we get mixed results for other languages and even performance drops for high-resource languages.
%we propose computationally cheap methods for post-hoc improving of . 

%4) We validate our benchmark and analyse the impact of the partial label translation on accuracy and we compare our benchmark to three image-text datasets and show that retrieval performance and accuracy on our benchmark generally correlate, though there are model-specific idiosyncracies that we discuss.

% PLACED HERE so it is on top of §3
\begin{figure*}[t!]
    \centering
    \includegraphics[width=0.8\textwidth]{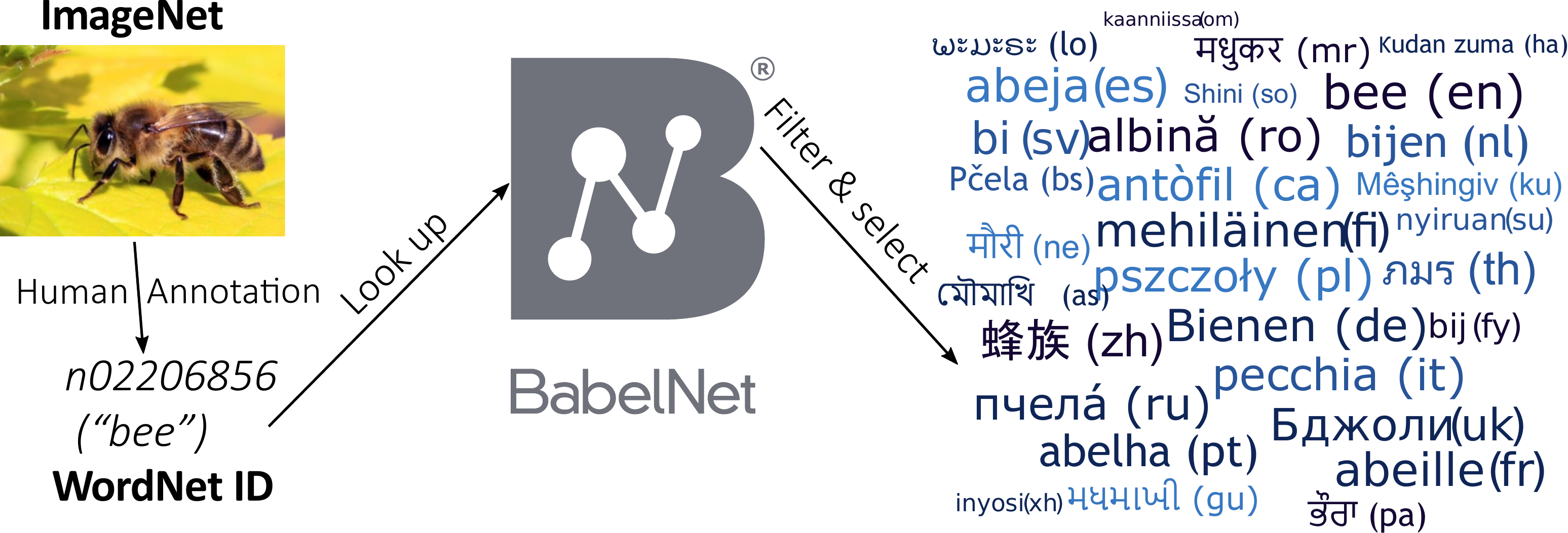}
    \caption{Illustrating the creation of Babel-ImageNet: ImageNet classes correspond to WordNet IDs, which are integrated into BabelNet, a multilingual semantic net. Through this, we look up synonymous word senses in all available languages, perform some cleaning and filtering and select one sense as label. }
    \label{fig:creation}
\end{figure*}

\section{Related Work}

% We first provide an overview of existing benchmarks for evaluating multilingual VL models, followed by a brief overview of multilingual CLIP models, commonly used for efficient image-text retrieval. 

% \subsection{Multilingual Vision-and-Language Benchmarks}
\paragraph{Multilingual Vision-and-Language Benchmarks.}
Early multilingual VL models \citep{gella_image_2017,wehrmann_language-agnostic_2019,kim_mule_2020,burns_learning_2020,huang_m3p_2020,geigle_retrieve_2022,zhou_uc2_2021} were often evaluated in image-text retrieval on Multi30k \citep{elliott_multi30k_2016,elliott_findings_2017,barrault_findings_2018}, an extension of Flickr30k \citep{plummer_flickr30k_2015} to German, French, and Czech, as well as on the Japanese \citep{yoshikawa_stair_2017} and Chinese \citep{li_coco-cn_2019} translations of MSCOCO \citep{lin_microsoft_2014}. More recent models were evaluated on multilingual image-text retrieval benchmarks: XTD \citep{aggarwal_towards_2020} (10 languages) and WIT \citep{srinivasan_wit_2021} (108 languages). Both these benchmarks, however, have prominent shortcomings. XTD predominantly contains examples from Karpathy's training portion of MSCOCO \citep{karpathy_deep_2017}, which is commonly used for pretraining of VL models, which constitutes a case of data leakage \citep{bugliarello_iglue_2022}.
% : this constitutes a case of data leakage because XTD's multilingual captions are directly translated from the original English captions \citep{bugliarello_iglue_2022}. 
WIT collects image-caption pairs from Wikipedia(s), which are abundant with named entity mentions that are often identical across a number languages  -- this artificially equates the difficulty of retrieval across languages \citep{zhai_lit_2022}.
 % (2) The number of image-caption pairs for a language depends on the size of its Wikipedia: this not only prevents direct performance comparisons across languages but also results in overly optimistic performance estimates for low-resource languages with merely a few hundred image-caption test pairs.
More recently, IGLUE \citep{bugliarello_iglue_2022} was introduced as the first benchmark to also include reasoning tasks like visual QA \citep{pfeiffer_xgqa_2021}, primarily meant to test cross-encoder models that jointly encode image-text pairs \citep{huang_m3p_2020,zhou_uc2_2021,zeng_cross-view_2022}.
IGLUE also introduces the image-caption dataset xFlickrCo, a combination of Flickr30k and MSCOCO with new captions in 7 languages.  
Another recent dataset, XM3600 \citep{thapliyal_crossmodal-3600_2022}, encompasses 3600 images (balanced by geography of origin) with captions in 36 languages written from scratch by native speakers. 
%Despite high-quality\footnote{The authors explicitly acknowledge the very high cost of human captioning of images in 36 languages (regretfully, they do not explicitly disclose the cost).}, it has yet to become widely adopted -- along with \citet{zhai_sigmoid_2023}, we are among the first to use it.

Motivated by monolingual CLIP models in other languages, translations of ImageNet classes have emerged for a handful of high-resource languages: Italian (obtained with MT) \citep{bianchi_contrastive_2021}, Chinese (human translations) \citep{yang_chinese_2022}, and Japanese\footnote{\href{https://github.com/rinnakk/japanese-clip}{github.com/rinnakk/japanese-clip}} and Arabic\footnote{\href{https://github.com/LAION-AI/CLIP_benchmark/pull/68}{github.com/LAION-AI/CLIP\_benchmark/pull/68}} (undisclosed methods). Extending ImageNet to more languages, however, is not feasible at scale, 
due to the challenges of finding native speakers to translate or verify machine-translated labels.
% : on the one hand, manual translation requires finding native speakers of low-resource languages (additionally fluent in English), which is challenging; machine translation of concepts (i.e., words and short phrases) out of context, on the other hand, is problematic due to polysemy -- it thus also requires validation of the translated senses by bilingual annotators: this is especially critical for low-resource languages for which current MT systems are still lacking \citep{costa-jussa_no_2022}.
%
With \bin{}, we exploit the linkage between ImageNet and BabelNet through WordNet, 
% In contrast, by exploiting the fact that ImageNet classes correspond to Wordnet synsets which, in turn, have corresponding massively multilingual synsets in BabelNet, we are able 
to create the first robust massively multilingual translation of ImageNet classes, avoiding the caveats of polysemy associated with automatic translation of concepts. 
%\bin{} allows for the evaluation of more languages than ever before directly on ImageNet, aligning multilingual CLIP models to the standard evaluation of their English counterparts. 

% \subsection{Multilingual CLIP}
\paragraph{Multilingual CLIP.}
CLIP \cite{radford_learning_2021} is not the first model to embed images and text in a shared representation space, but it has arguably become the most widely used one, owing its effectiveness -- especially in ZS-IC -- to the immense pretraining corpus. Older models, not exposed to large-scale VL pretraining, e.g., \citep{faghri_vse_2018} for English and \citep{gella_image_2017,wehrmann_language-agnostic_2019,kim_mule_2020,burns_learning_2020} multilingually, focused predominantly on text-to-image retrieval and were not shown to exhibit ZS-IC abilities. 
%and earlier work also did not use large-scale multimodal pretraining as CLIP did.
%%
MURAL \citep{jain_mural_2021} was the first -- albeit not publicly released -- multilingual CLIP model, trained on billions of multilingual image-caption pairs. 
% To the best of our knowledge, 
The only publicly available multilingual CLIP models trained ``from scratch'' are the OpenCLIP models \citep{ilharco_openclip_2021} (trained on the full multilingual LAION5B dataset \citep{schuhmann_laion-5b_2022})
% -- pretrained using the full multilingual LAION5B dataset \citep{schuhmann_laion-5b_2022}, consisting of 5B image-caption pairs covering 100+ languages.
and mSigLIP \cite{zhai_sigmoid_2023} (trained on Google's WebLI \cite{chen_pali_2022}).
Monolingual CLIP models for a few languages other than English (e.g., Italian, Chinese) have also been released \citep{bianchi_contrastive_2021,yang_chinese_2022}: due to comparatively small pretraining data, they trail the English models. 
%and compute compared to the English CLIP models, they do not match up in performance.
Given the huge computational cost of training a multilingual CLIP from scratch, teacher distillation \citep{reimers_making_2020} has become popular as an efficient alternative \citep{carlsson_cross-lingual_2022,chen_altclip_2022,zhang_generalizing_2022}: a multilingual text encoder (e.g., XLM-R \cite{conneau_unsupervised_2020}) is forced (commonly using parallel sentences) to align its representation space to the text encoder of English CLIP.

% \section{\bin: Massively Multilingual Zero-Shot Image Classification}

\section{\bin}
\label{sec:dataset}

\paragraph{Why (massively) multilingual ZS-IC?} With class labels in a particular language we can evaluate VL models in language-specific ZS-IC. Note that the goal is not to improve the classification performance -- using labels in any other language yields worse performance compared to using English labels. Instead, we argue that a model's language-specific ZS-IC performance is a good estimate of the quality of its multilingual VL representation space for the language, and thus a good predictor of the model's performance for that language.
%in ``real'' tasks (e.g., image-caption retrieval).
%for which massively multilingual data collection is prohibitively expensive. 
%we demonstrate that this is the case for image-caption retrieval (see \S\ref{sec:experiments:retrieval}) that it correlates with language-specific image-caption retrieval performance.

In addition, prior work evaluated models mainly with retrieval. 
For the languages, where such data exists, we provide a more comprehensive evaluation: ImageNet covers a far more diverse set of concepts than image captions usually contain. 

\paragraph{WordNet as a matchmaker for ImageNet and BabelNet.} Unlike in most image classification datasets (e.g., CIFAR10, Oxford Pets \citep{parkhi_cats_2012}, Flowers102 \citep{nilsback_automated_2008}), where image classes are \textit{words}, ImageNet \citep{deng_imagenet_2009} links images to \textit{concepts}, represented with sets of synonyms (synsets) from English WordNet \citep{miller_wordnet_1994}.
BabelNet \citep{navigli_babelnet_2010} is a massively multilingual lexico-semantic network, automatically created by merging and consolidating numerous lexico-semantic resources: from WordNets in dozens of languages (e.g., \cite{hamp1997germanet,pianta2002multiwordnet}) to (massively multilingual) Wikipedia and WikiData \cite{vrandevcic2012wikidata}.\footnote{BabelNet v5.2 consolidates 53 sources 
% (\href{https://babelnet.org/statistics}{source})
} Crucially for our efforts, BabelNet is (1) also organized in (multilingual) synsets, containing synonyms across many languages and (2) each of its synsets has an explicit link to the corresponding (English) WordNet synset (if such exists). 
With WordNet as the seam between ImageNet and BabelNet, we are able to create a massively multilingual ZS-IC benchmark, without resorting to manual annotation or MT.

\paragraph{Class Translation and Cleaning Process.} 
We illustrate the general process in Figure~\ref{fig:creation}.
For every ImageNet class, using its WordNet synset ID, we fetch all synonyms in every language including English (where available). 
Next, we remove words that match an English word (of the same class) because models use their high-quality English representations for these words, distorting results and leading to misleadingly optimistic estimates of models' multilingual abilities.
Next, we eliminate all words that were added to BabelNet via machine translation, removing the potential negative effects of context-agnostic MT from our benchmark.   
Finally, we select for every class in every language the first remaining word (according to the order in BabelNet) as our final language-specific class label. 
Classes with no word are removed.

% For each ImageNet class, we first fetch all words of each synset in all languages (where available) including English from the corresponding BabelNet synset (using the WordNet synset ID for matching).  
% %%
% We next remove words from other languages for which an identical English word exists. We do this because having target language labels identical to English labels would allow multilingual VL models to rely on their high-quality English representations. With those being semantically (much) better than representations of other languages, this would lead to misleadingly optimistic estimates of models' multilingual abilities. On average, this removes $84 \pm 39$ classes from a language-specific benchmark (i.e., we lose classes for which all available BabelNet words in a language also exist as English words in the same synset).
% %%%    
% Next, we eliminate all words that were added to BabelNet via machine translation, removing the potential negative effects of context-agnostic MT from our benchmark.   
% This mostly affects high(er)-resource languages and removes on average $148 \pm 184$ classes.
% %%%
% Finally, we select for every remaining class the first words in the respective language (according to the order in BabelNet) as our final language-specific class label. 

\paragraph{Language Selection.}
We find that of the 520 languages in BabelNet v5.2, 298 have at least 10 classes using our process, the majority of them being low- and very-low resource languages.
We limit our evaluation to 100 non-English languages for more balanced selection of low-, mid- and high-resource languages.
To this end, we combine the 92 languages (counting unique ISO codes) covered by the pretraining corpora of XLM-R \citep{conneau_unsupervised_2020} with 8 manually selected languages not covered by machine translation (neither Google, Microsoft, nor by NLLB \cite{costa-jussa_no_2022}).

% While one could obtain ImageNet class translations in all BabelNet languages,\footnote{BabelNet v5.2 covers 520 languages of which 298 have at least 10 synsets that correspond to classes of ImageNet; we release labels for these 298 languages along with our code so that anyone with access to BabelNet can create translations in additional languages.} we limit our evaluation to 92 non-English languages (counting unique ISO codes) covered by the pretraining corpora of XLM-R \citep{conneau_unsupervised_2020}.
% The motivation for this decision is twofold: (1) 50K images in ImageNet and up to 80K label + prompt combinations per language -- to be embedded with each model included in our comparative evaluation -- make even the ZS-IC evaluation computationally intensive, limiting the total number of languages for which we could carry it out with our computational resources; (2) the majority of publicly available multilingual CLIP models have been derived precisely from XLM-R as the initial multilingual text encoder.

\paragraph{Grouping Languages in Evaluation.}
Comparing models' performance over 100 languages (+English) is still unwieldy but averaging performance across all languages is too reductive and consequently not particularly informative. 
We thus opt for the middle ground: we group \bin{} languages in four buckets based on their number of classes ($<$101, 101 to 333, 334 to 667, and 668 to 1000). We argue that the number of classes is a reasonable proxy for general ``resourceness'' of a language (see \S\ref{sec:appendix:detail:data} for the full list of languages and corresponding numbers of classes) and accordingly designate the four groups as \textit{very-low-}, \textit{low-}, \textit{mid-}, and \textit{high-resource}, encompassing 17, 32, 35, and 16 languages, respectively.
Additionally, given that ZS-IC becomes easier with fewer classes, averaging results across languages with more comparable numbers of classes makes more sense than averaging them across all languages. 
% Nonetheless, due to differing sets of classes, we caution against direct performance comparisons of results \textit{between} groups or across individual languages. Instead, for any particular language or language group \bin{} allows for a direct comparison of competing multilingual VL models.    

\paragraph{Verification.}
BabelNet mappings are automatically created and not error-free even though, for example, over 90\% of Wikipedia-WordNet mappings were manually validated with a precision over 99.5\% \cite{navigali_ten_2021}.
Still, we also manually verify our mappings on 4 languages with native speakers\footnote{The authors and colleagues. Languages: \textit{de, ur, tr, hr}} and on the 3 smallest languages (\textit{xh, om, ha}) with online dictionaries; we find on average 5.4\% errors.
We believe this to be a very acceptable error rate, considering (i) that around 6\% of ImageNet images are mislabeled \citep{northcutt_pervasive_2021} and (ii) that there are also erroneous mappings between ImageNet images and WordNet synsets \citep{nielsen_linking_2018,radford_learning_2021}.

\section{Benchmarking CLIP Models}

\begin{table}[]
    \centering
    \small
    \footnotesize
     \def\arraystretch{0.97}
     \resizebox{1\linewidth}{!}{
    \begin{tabular}{llrr}
    \toprule
    \bf Source & \bf Objective & \bf \#Data & \bf \#Langs \\
    \midrule
\texttt{OpenAI} & CLIP & 400M & 1 \\
\texttt{OpenCLIP} & CLIP/LiT & 5B  & $>$100 \\
\texttt{M-CLIP} & distill & 3M  & 69 \\
\texttt{M-CLIP}  & distill & 7M & 48 \\
%M-CLIP & LaBSE & L-14 & distill & 7m & 48 \\
\texttt{SentenceTransformer}  & distill & $>$50M  & 49 \\
\texttt{AltCLIP} & distill+LiT & 50M+100M & 9 \\
\texttt{SigLIP} & CLIP & 900M & 100 \\
\texttt{NLLB-SigLIP} & LiT & 20M & 201 \\ 
    \bottomrule
    \end{tabular}
    }
    \caption{CLIP variants benchmarked with: (i) the source (who trained the model), (ii) the training objective (CLIP: contrastive training as in \citet{radford_learning_2021}, LiT: locked image tuning \citet{zhai_lit_2022}, distill: MSE teacher distillation \citep{reimers_making_2020}), (iii) training data (for ``distill'' the number of caption pairs, for CLIP/LiT the number of image-text pairs), and (iv) the number of languages seen in training.
    }
    
    \label{tbl:models}
\end{table}

\begin{table*}[t]
% \begin{table}[t]
    \centering
    \footnotesize
     % \def\arraystretch{0.97}
     % \resizebox{0.99\linewidth}{!}{
    \begin{tabular}{l rrrrrr}
    \toprule
\bf Model & \bf Param. & \bf v-low & \bf low & \bf mid & \bf high & \bf en \\
\midrule
%        OpenAI B-32 &      4.2 &      4.9 &      9.0 &     61.3 \\
% OpenCLIP XLMR B-32 &     15.0 &     31.0 &     \textbf{39.7} &     \textbf{62.8} \\
%   M-CLIP XLMR-L B-32 &     25.7 &     32.8 &     33.3 &     42.6 \\
%  M-CLIP XLMR-L B-16+ &     \textbf{25.8} &     \textbf{34.5} &     36.0 &     46.4 \\
%  M-CLIP mBERT B-32 &     14.8 &     19.3 &     18.9 &     29.2 \\
%      ST mBERT B-32 &      9.2 &     15.1 &     17.1 &     38.2 \\
%            \cmidrule(lr){2-5} 
%   M-CLIP XLMR-L L-14 &     \textbf{28.1} &     37.7 &     39.5 &     51.6 \\
%  AltCLIP XLMR-L L-14 &     14.2 &     21.1 &     33.6 &     69.9 \\
% OpenCLIP XLMR-L H-14 &     19.5 &     \textbf{41.1} &     \textbf{52.4} &     \textbf{77.1} \\
         OpenAI B-32 &   151m &       4.28 &      3.79 &      5.02 &       9.23 & 63.35 \\
       ST mBERT B-32 &   286m &       6.23 &      9.72 &     15.33 &      17.44 & 39.45 \\
   M-CLIP mBERT B-32 &   330m &      10.16 &     15.42 &     19.63 &      19.26 & 29.97 \\
OpenCLIP XLMR-B B-32 &   366m &      12.00 &     18.29 &     30.86 &      39.52 & 62.32 \\
             mSigLIP &   371m &      17.33 &     29.05 &     \textbf{48.20} &      \textbf{56.66} & \textbf{75.12} \\
    NLLB-SigLIP-base &   507m &      \textbf{34.11} &     \textbf{34.58} &     32.17 &      29.37 & 39.75 \\
  M-CLIP XLMR-L B-32 &   712m &      18.52 &     26.40 &     33.47 &      34.11 & 44.06 \\
 M-CLIP XLMR-L B-16+ &   769m &      18.92 &     27.62 &     34.98 &      36.46 & 47.02 \\
 \midrule
 AltCLIP XLMR-L L-14 &   864m &      12.67 &     16.98 &     21.32 &      33.97 & 73.36 \\
  M-CLIP XLMR-L L-14 &   988m &      19.80 &     29.70 &     38.17 &      40.07 & 52.34 \\
OpenCLIP XLMR-L H-14 &  1193m &      13.77 &     23.57 &     41.03 &      \textbf{52.23} & \textbf{76.95} \\
   NLLB-SigLIP-large &  1195m &      \textbf{40.61} &     \textbf{43.22} &     \textbf{42.78} &      39.75 & 51.96 \\
    \bottomrule
    \end{tabular}
     % }
    \caption{ZS-IC performance on \bin{}: average results for very-low-/low-/mid-/high-resource languages and English.  \textbf{Bold}: best result in each column, both between models with base (B) and large (L/H) image encoders.
}
    \label{tbl:results:main}
% \end{table}
\end{table*}

%We next detail our experimental setup and benchmark a range of publicly available multilingual CLIP models on on \bin.

\textbf{Models.} We present the CLIP variants we benchmark on \bin{} (overview in Table\,\ref{tbl:models}). 
All use Vision Transformers (ViT) \citep{dosovitskiy_image_2020} albeit of different sizes and with differently sized input patches (e.g., B-32 = \texttt{Base} Transformer with 32$\times$32-pixel patches).

\texttt{OpenAI:} The original CLIP \citep{radford_learning_2021}, trained fully from scratch on 400M English image-caption pairs. 
%The text encoder is trained from scratch.
%(i.e., not initialized with any pretrained weights).

\texttt{OpenCLIP:} OpenCLIP \citep{ilharco_openclip_2021} aims to replicate the OpenAI models using the public LAION datasets \citep{schuhmann_laion-400m_2021,schuhmann_laion-5b_2022}.
Two multilingual models have been trained: the B-32 model with the text encoder initialized with the weights of XLM-R-Base and the H-14 model initialized with XLM-R-Large.
The B-32 variant is trained with the original contrastive CLIP objective, whereas the H-14 model was trained via locked image tuning (LiT, \citep{zhai_lit_2022}) in which the pretrained image encoder of the English H-14 OpenCLIP model is frozen and only the parameters of the text encoder are updated. 

\texttt{SentenceTransformer\,(ST):} One of the first distillation-based multilingual CLIP-like models using the approach of \citet{reimers_making_2020},\footnote{\href{https://huggingface.co/sentence-transformers/clip-ViT-B-32-multilingual-v1}{\textsc{clip-ViT-B-32-multilingual-v1}}} with over 50M EN-X parallel sentences (X being one of 49 other languages) as supervision. They distill a (distilled) mBERT student \citep{devlin_bert_2019} from the English OpenAI B-32 teacher.
%. as the teacher with 50+ million aligned English-X sentence pairs in 49 languages as training data.

\texttt{M-CLIP:} Multilingual-CLIP (M-CLIP) \cite{carlsson_cross-lingual_2022} is a model distilled using mBERT and OpenAI B-32 with translations of 3M English image captions 
%(from public image-text datasets) 
to 69 languages as parallel supervision. 
Post-publication, they released a set of models\footnote{\href{https://github.com/FreddeFrallan/Multilingual-CLIP/blob/main/larger_mclip.md}{github.com/FreddeFrallan/Multilingual-CLIP}} with XLM-R-Large as the student and  OpenAI B-32, L-14, and OpenCLIP B-16+ as teachers. These models were trained on 7M captions by \citet{li_blip_2022}, translated to 48 languages.
% The original English captions were \textit{not} used in training.

\texttt{AltCLIP:} This model by \citet{chen_altclip_2022} distills an XLM-R-Large student with OpenAI L-14 as teacher, targeting 9 languages and using as training data a mix of machine-translated captions, multilingual captions sampled from LAION5B, and aligned English-X sentence pairs. After distillation, the authors additionally fine-tune the model via LiT using selected image-text pairs from LAION5B in the 9 target languages.

\texttt{SigLIP}: The multilingual mSigLIP \cite{zhai_sigmoid_2023} is a B-16-size CLIP model trained from scratch on Google's multilingual WebLI dataset \cite{chen_pali_2022}. They replace the softmax function used in the original CLIP with a sigmoid for more compute-efficient training.

\texttt{NLLB-SigLIP}: The NLLB-CLIP models \cite{visheratin_nllb_2023} are a suite of CLIP models trained with LiT using 100k image-caption pairs translated fully for all 200 languages supported by NLLB \cite{costa-jussa_no_2022} for 20M examples in total.
They use SigLIP B-16 and SO400M-14 as image encoders with the encoders of NLLB-600M-distilled and NLLB-1.3B-distilled as text encoders, which we call \textit{base} and \textit{large}, respectively.

\paragraph{Zero-Shot Image Classification Setup.}
%\label{sec:experiments:setup}
We adopt the ZS-IC setup of \citet{radford_learning_2021}: for an image-label pair, the image embedding is obtained directly from the image encoder; the label is inserted into 80 different prompt templates (from \cite{radford_learning_2021}), each of which is independently embedded by the text encoder -- the final label representation is then the mean of prompt embeddings.
%an embedding of a class label is obtained by averaging the representations  and textual class labels (prepended with 80 task-specific prompts) are encoded by the respective image and text tower.
The class with the label embedding that is most similar to the image embedding (according to cosine similarity) is taken as the prediction; accuracy (top-1) is the evaluation metric.
%%
%In addition to results for the low/ mid/ high-resource groups, we also report English results using the class labels and prompts by \citep{radford_learning_2021}.

\noindent\textit{Translating Prompts.}
We translate the 80 English prompts used by \citet{radford_learning_2021} to our 100 languages using NLLB \citep{costa-jussa_no_2022} (\texttt{nllb-200-distilled-1.3B}; see \S\ref{sec:appendix:detail:prompts} for details). 
We show (see \S\ref{sec:experiments:prompts}) that translated, language-specific prompts lead to better performance compared to using only the class labels or inserting them into the original English prompts. Moreover (see \S\ref{sec:experiments:mt_imagenet}), we show that translated prompts yield similar performance as human-crafted prompts in \textit{ar}, \textit{it}, \textit{ja}, and \textit{zh}.

\paragraph{ZS-IC Results.} Table \ref{tbl:results:main} summarizes the results for the four language groups, alongside the English performance. The full results for all 100 languages can be found in the Appendix (Table \ref{tab:res:full}). 

% Although differing subsets of ImageNet classes across language-specific benchmarks (see \S\ref{sec:dataset}) prevent direct comparison of numbers between the language groups, 
These results make it abundantly clear that multilingual CLIP models perform dramatically worse (i) for high-resource languages than for English, and (ii) for low- and mid-resource languages than for high-resource languages. Note that this is \textit{despite} the classification tasks \textit{a priori} being easiest for low-resource languages (under 333 classes) and hardest for English, where models must distinguish between all 1000 classes of ImageNet.

The English ImageNet performance of the models is \textit{not} indicative of their ZS-IC performance for other languages, especially low-resource ones: for example, OpenCLIP XLMR-L H-14 outperforms M-CLIP XLMR-L L-14 by 25 accuracy points on English ImageNet, yet trails it 8.6 points on average for low-resource languages. We believe that this points to the ``curse of multilinguality'' of the text encoder -- namely that, under a fixed model capacity, an improvement of representation quality for some language(s) comes at the expense of representational deterioration for others. This phenomenon has been well-documented in particular for XLM-R \cite{conneau_unsupervised_2020,pfeiffer_mad-x_2020}.     
Among the model variants obtained with the same training procedure (e.g., four variants of M-CLIP), English performance does seem to correlate with the performance on other languages. 

The OpenCLIP models and mSigLIP, both trained on massive web-crawled corpora, yield good results for high- and mid-resource languages but perform poorly (in comparison to M-CLIP and NLLB-SigLIP) for low-resource languages.
%, with mSigLIP still performing relatively better despite its small parameter count. 
In \S\ref{sec:experiments:lang_distribution}, we demonstrate that  OpenCLIP performance strongly correlates with the distribution of languages in LAION5B: this would suggest that contrastive training (i.e., CLIP and LiT) leads to poor cross-language generalization. 

In contrast, the better performance of (XLM-R-based) M-CLIP models on low-resource languages suggests that distillation-based training offers stronger cross-lingual generalization (and yields best performance even for languages unseen in distillation training, see \S\ref{sec:experiments:distillation}). 
We hypothesize that by aligning representations of captions in all other languages to the representations of corresponding English captions results in a more language-agnostic representation space. At the same time, in line with the ``curse of multilinguality'', this improved generalization is paid with reduced quality of representations of high-resource languages, where M-CLIP models fall well behind. 

The NLLB-SigLIP models, which combine a strong image encoder with the NLLB encoders trained on 200 languages, show impressive results for low- to mid-resource languages but underperform for the high-resource languages, suggesting again a trade-off between languages.

The trade-off between cross-lingual generalization and per-language performance is best exemplified with AltCLIP: the model is exceptionally good for the 9 languages present in its large-scale distillation training (\S\ref{sec:experiments:distillation}), yet performs (comparatively) poorly for most other languages -- training on a very large dataset for only a few languages simply overwrites the XLM-R's knowledge of other languages, obtained in its original pretraining. 

The two mBERT-based models significantly underperform all other models. This is in part due to mBERT being generally a weaker multilingual text encoder \citep{hu2020xtreme,lauscher2020zero}. On top of that, M-CLIP mBERT variants have been trained on less data than XLM-R-based counterparts (3M vs. 7M captions) and ST is distilled with parallel sentences that are \textit{not} image captions.

\section{Validating \bin{}}
\label{sec:validation}
%%%

\begin{figure}
    \centering
    \includegraphics[width=0.8\linewidth]{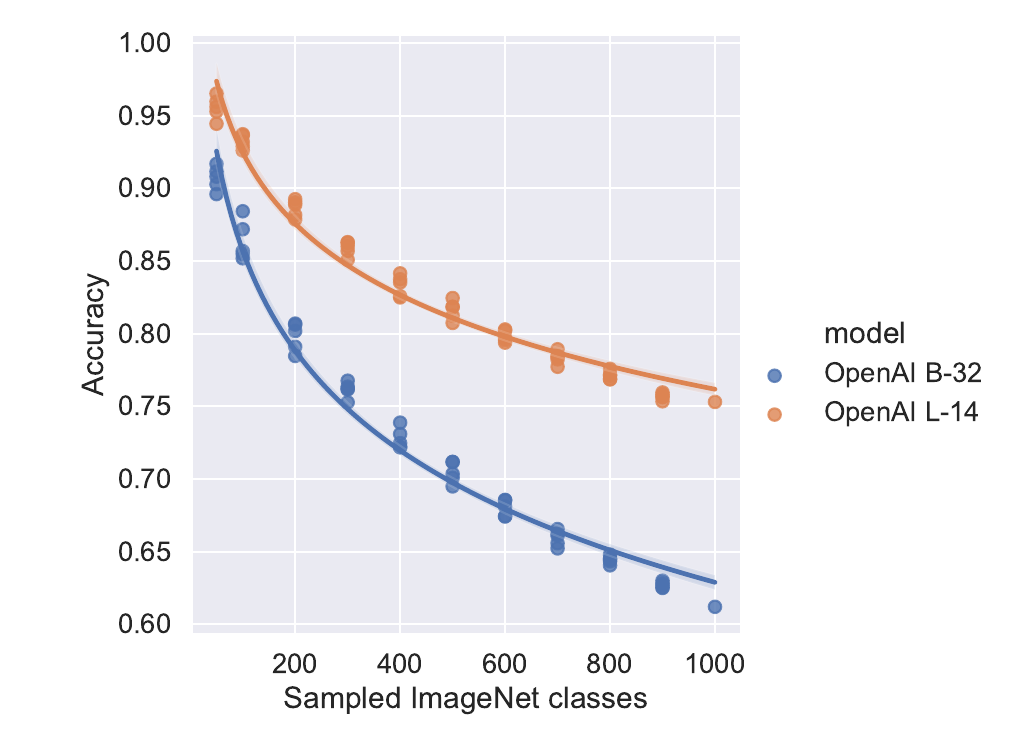}
  \caption{English ImageNet results with random subset of the 1k classes (5 random seeds each).
  }
\label{fig:plot:less1k_analysis}
\end{figure}

\begin{table}
\centering
\footnotesize
     \def\arraystretch{0.97}
     \resizebox{0.99\linewidth}{!}{
\begin{tabular}{lrrrrr}
\toprule
                \bf Model & \bf v-low & \bf low & \bf  mid & \bf  high & \bf  en \\
\midrule
%         \texttt{OpenAI} B-32 &      5.0 &      7.7 &     15.5 &     84.1 \\
%  \texttt{OpenCLIP} XLMR B-32 &     16.4 &     39.5 &     \textbf{56.2} &     \textbf{85.3} \\
%   \texttt{M-CLIP} XLMR-L B-32 &     26.1 &     41.2 &     47.3 &     66.3 \\
%  \texttt{M-CLIP} XLMR-L B-16+ &     \textbf{26.6} &     \textbf{42.2} &     48.6 &     70.7 \\
%   \texttt{M-CLIP} mBERT B-32 &     15.9 &     26.5 &     30.3 &     49.9 \\
%       ST mBERT B-32 &     10.4 &     22.3 &     28.6 &     60.1 \\
% \midrule
%   \texttt{M-CLIP} XLMR-L L-14 &     \textbf{28.2} &     45.2 &     51.9 &     72.2 \\
%  \texttt{AltCLIP} XLMR-L L-14 &     15.7 &     27.5 &     45.7 &     90.9 \\
% \texttt{OpenCLIP} XLMR-L H-14 &     20.0 &     \textbf{48.7} &     \textbf{66.5} &     \textbf{92.6} \\
       ST mBERT B-32 &       6.23 &     11.83 &     24.36 &      32.65 & 64.79 \\
   M-CLIP mBERT B-32 &      10.16 &     18.30 &     29.86 &      34.49 & 54.21 \\
OpenCLIP XLMR-B B-32 &      12.00 &     20.84 &     42.15 &      60.20 & 87.56 \\
             mSigLIP &      17.33 &     32.20 &     \textbf{59.83} &      \textbf{75.83} & \textbf{93.58} \\
    NLLB-SigLIP-base &      \textbf{34.11} &     \textbf{38.93} &     45.53 &      47.78 & 63.18 \\
  M-CLIP XLMR-L B-32 &      18.52 &     30.10 &     45.36 &      53.21 & 68.70 \\
 M-CLIP XLMR-L B-16+ &      18.92 &     30.91 &     45.75 &      54.14 & 70.31 \\
 \midrule
 AltCLIP XLMR-L L-14 &      12.68 &     19.42 &     29.20 &      49.49 & 93.35 \\
  M-CLIP XLMR-L L-14 &      19.80 &     32.99 &     49.43 &      58.13 & 75.02 \\
OpenCLIP XLMR-L H-14 &      13.77 &     26.32 &     52.65 &      \textbf{71.65} & \textbf{94.09} \\
   NLLB-SigLIP-large &      \textbf{40.61} &     \textbf{47.55} &     \textbf{56.25} &      58.84 & 74.59 \\

\bottomrule
\end{tabular}
}
\caption{ZS-IC results with the same number of classes for each language (we report averages over 10 random subsets of 100 classes per language). 
%For the 
%12
%17 very-low-resource languages with $<$100 classes, we take the single result on all classes.
}
\label{tbl:results:100classes}
\end{table}

\begin{figure*}[t]
    \centering
    \includegraphics[width=\linewidth]{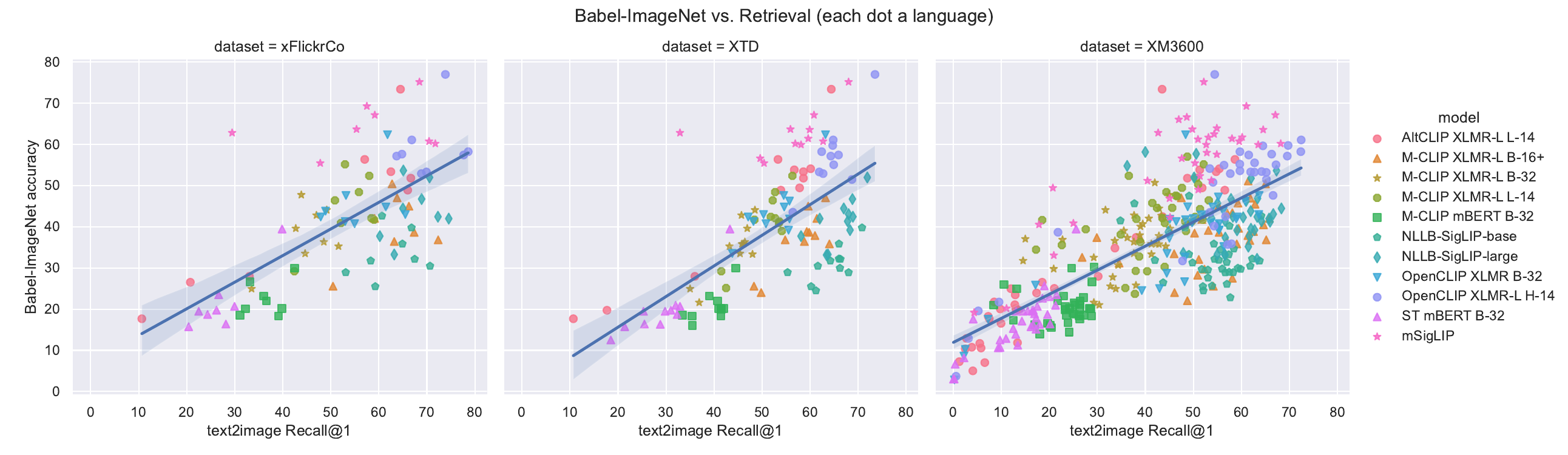}
    \caption{R@1 text-to-image retrieval results on three datasets plotted against \bin{} performance  (each dot denotes the performance of one model for one language) together with a linear regression estimate (95\% CI). 
    }
    %\vspace{-1mm}
\label{fig:plot:retrieval_classification}
%\vspace{-3mm}
\end{figure*}

We perform two additional analyses that establish the validity of \bin{} as a benchmark: (1) how different number of classes affects performance and findings across languages and (2) how ZS-IC performance on \bin{} relates to multilingual image-text retrieval performance. We provide further analyses in the Appendix. 
\label{sec:experiments:class_number}

\paragraph{Effect of number of classes on ZS-IC accuracy.} \bin{} is an incomplete translation of the ImageNet classes (see \S\ref{sec:dataset}). Intuitively, classification with fewer classes is easier and results in higher absolute performance for all models. 
%%%
Our analysis of how the number of classes affects the ZS-IC performance on the English ImageNet for OpenAI CLIP models (B-32 and L-14) confirms this. Figure \ref{fig:plot:less1k_analysis} shows that the task difficulty (i.e., ZS-IC performance) is log-linear in the number of classes: this makes intuitive sense -- moving from 50 to 100 classes increases the difficulty more than going from 900 to 950 classes. 

% The actual steepness of the estimate is model-dependent so there is not a single coefficient that we could use to adjust our results to the "real 1k " difficulty.
% Still, using the estimated $\alpha=-0.16$ of the L14 model to adjust scores, we see that our "overconfident" results for low-resource languages disappear: For the overall best H14, 27\% of languages (and 82\% of low-resource ones) achieve 0\% accuracy; for M-CLIP L14, which is best for low-resource, still 15\% (or 71\% of low-resource) would fail completely. 

We next fix the number of classes to 100 for all \bin{} languages (except for languages with $<$100 classes, for which we make no changes) and report the performance in Table \ref{tbl:results:100classes} (for each language, we average the results over 10 different randomly selected subsets of 100 classes).
% \footnote{It is not possible to select the exact same set of classes across all \bin{} languages because only one ImageNet-1k class  (\textit{bee}, Figure \ref{fig:dataset_illustration}) has BabelNet translations in all 92 languages.} 
While in absolute terms the ZS-IC performance increases compared to full class sets (Table \ref{tbl:results:main}), and gaps between the language groups widen
%(especially between mid- and high-resource, and English to all three)
, our observations do not change: NLLB-SigLIP still exhibits the best performance for low-resource languages, whereas OpenCLIP and mSigLIP are still best for high-resource languages. This renders the (full) \bin{} a reliable benchmark for directly comparing multilingual VL models. 

\paragraph{Multilingual ZS-IC \& multilingual image-text retrieval.} 
The existing body of work commonly evaluates multilingual VL models in image-text retrieval. 
One goal of \bin{}, which measures multilingual ZS-IC performance, is to reflect (or, at least, give an estimate of) the quality of the multilingual embedding spaces of VL model. 
% Proving that this is the case implies showing that the models' ZS-IC performance on \bin{} is indicative of their multilingual performance on tasks that -- unlike ZS-IC itself -- make sense in a multilingual formulation. 
%%
We thus compare how models' performance on \bin{} correlates with their performance on three different multilingual image-text retrieval datasets: xFlickrCo \citep{bugliarello_iglue_2022}, XTD \citep{aggarwal_towards_2020}, and XM3600 \citep{thapliyal_crossmodal-3600_2022}, covering 7, 10, and 34\footnote{For correlation analysis, we exclude \textit{mi} and \textit{quz} as they are not under the 100 \bin{} languages.
%we still report the models' image-text retrieval performance for those languages in Table \ref{tab:res:xm3600}
} languages.
%, 3600 images and $\approx$2 caption per image.
We use R@1 in text-to-image retrieval as the evaluation metric: it captures the percentage of examples where a correct image is top-ranked for a given caption. We report the full retrieval results in the Appendix (Tables \ref{tab:res:xflickrco}, \ref{tab:res:xm3600}, and \ref{tab:res:xtd}).
      
%and accuracy on our benchmark correlate (with some model-specific idiosyncracies), which makes our benchmark suitable for large-scale evaluation of multilingual multimodal representations.

% \begin{figure*}[]
%     \centering
%     \includegraphics[width=\linewidth]{figures/plots/retrieval_imagenet_mean_correlation.pdf}
%     \caption{Average T2I retrieval results plotted against Babel-ImageNet. 
%     }
%     %\vspace{-1mm}
% \label{fig:plot:retrieval_classification_average}
% %\vspace{-3mm}
% \end{figure*}

Figure \ref{fig:plot:retrieval_classification} displays the retrieval results on xFlickrCo, XM3600, and XTD, respectively, against the ZS-IC accuracy on \bin{}: each dot represents one model-language combination.
The plots reveal high correlation between the \bin{} and text-to-image retrieval scores across model-language pairs: $0.67$ for xFlickrCo, $0.75$ for XM3600, and $0.66$ for XTD. 
It is particularly positive that \bin{} shows the highest correlation with XM3600, with which it intersects in most languages (34). 
These results confirm that \bin{} is a sensible benchmark for comparing proficiency of VL models for a multitude of languages not covered by existing benchmarks.
% in task-specific (e.g., image-text retrieval) benchmarks.  

The evaluation also reveals model-specific idiosyncrasies that consideration of either task alone would not have shown.
For example, models that use a higher image resolution (M-CLIP B-16+, NLLB-SigLIPs) perform relatively better for retrieval, where the increased detail helps, than for ZS-IC. 
mSigLIP, potentially due to its small text encoder (compared to, e.g., XLM-R-large), is not as strong for retrieval as in ZS-IC.
With the addition of ZS-IC to formerly retrieval-only multilingual evaluation, \bin{} allows for a more comprehensive evaluation of models.

% Balancing the number of classes to 100 for each language results in even higher correlations, e.g., $0.89$ NOT UPDATED VALUE against XM3600. 

%Interestingly, fixing the number of classes in the benchmark for comparable results between languages, as we do in \S\ref{sec:experiments:class_number}, does not improve the correlation much (i.e. by less than 0.03 points).
%
%, justifying again our decision of using all available classes for each language.

%%%%%%% GG: this can really be left out :)
%%%%
%Despite the overall correlation, the two tasks have their own idiosyncrasies and we notice that specific models can perform better on one of the two tasks: M-CLIP B16+, for example, works exceedingly well for retrieval---even outperforming the larger M-CLIP L14 model.
%%---and AltCLIP works noticeably better for classification than retrieval.
%We hypothesize that the larger image resolution for B16+ is helpful for retrieval but less so for classification where the to-be-classified objects take up a large part of the image anyway.
%%, and it is possible that the contrastive learning step after distillation for AltCLIP might help for classification.

\section{Improving Multilingual CLIP for Low-Resource Languages}
\label{sec:experiments:one_lang_train}

\begin{table*}[t]
    \centering
    \footnotesize
     \def\arraystretch{0.97}
     \resizebox{\linewidth}{!}{
    \begin{tabular}{ll rrrrrrrrrrrrrrrr}
    \toprule
 \bf Model & \bf Loss & \bf xh & \bf si & \bf lo & \bf ur & \bf my & \bf hi & \bf ms & \bf et & \bf sk & \bf lt & \bf eu & \bf ar & \bf ko & \bf fa & \bf de & \bf zh\\
\midrule
% Overall Best & No training & 27.8 & 36.3 & 18.0 & 37.2 & 22.4 & 41.1 & 48.5 & 48.1 & 58.8 & 45.7 & 21.4 & 41.7 & 53.2 & 42.6 & 61.2 & 53.5 \\
% \midrule
M-CLIP XLMR-L B-32 & No training & 17.7 & 33.6 & 12.5 & 29.4 & 14.6 & 36.4 & 36.6 & 41.4 & 39.7 & 27.5 & 18.3 & 30.1 & 21.4 & 25.0 & 38.7 & 32.7 \\
 \cmidrule(lr){2-2}
 & Text Contrastive & \cellcolor{blue!15}49.0 & \cellcolor{teal!15}46.1 & \cellcolor{teal!15}23.4 & \cellcolor{cyan!15}38.4 & \cellcolor{teal!15}33.3 & \cellcolor{orange!15}36.2 & \cellcolor{orange!15}34.0 & \cellcolor{red!15}34.6 & \cellcolor{orange!15}35.8 & 30.5 & \cellcolor{cyan!15}27.1 & \cellcolor{orange!15}25.3 & 23.4 & 26.1 & \cellcolor{red!15}31.4 & \cellcolor{orange!15}28.8 \\
 & LiT & \cellcolor{blue!15}44.5 & \cellcolor{teal!15}49.9 & \cellcolor{teal!15}24.7 & \cellcolor{teal!15}40.8 & \cellcolor{teal!15}29.4 & 37.3 & \cellcolor{orange!15}36.1 & \cellcolor{red!15}33.6 & \cellcolor{orange!15}35.2 & 31.2 & \cellcolor{teal!15}29.0 & \cellcolor{orange!15}25.9 & 24.7 & 27.0 & \cellcolor{red!15}33.6 & \cellcolor{orange!15}28.1 \\
 & MSE & \cellcolor{blue!15}46.8 & \cellcolor{teal!15}53.3 & \cellcolor{teal!15}26.9 & \cellcolor{teal!15}43.0 & \cellcolor{blue!15}37.2 & \cellcolor{cyan!15}42.5 & 39.3 & \cellcolor{orange!15}38.5 & 41.0 & \cellcolor{cyan!15}35.3 & \cellcolor{teal!15}34.8 & \cellcolor{orange!15}29.1 & \cellcolor{cyan!15}29.9 & \cellcolor{cyan!15}31.8 & \cellcolor{orange!15}38.1 & \cellcolor{orange!15}31.3 \\
 %\cmidrule(lr){2-2}
 %& MSE (WikiData) & --- & \cellcolor{red!15}25.0 & --- & --- & --- & \cellcolor{red!15}22.2 & --- & \cellcolor{red!15}21.3 & \cellcolor{red!15}23.3 & \cellcolor{red!15}21.8 & \cellcolor{orange!15}16.9 & \cellcolor{red!15}24.0 & \cellcolor{red!15}16.3 & \cellcolor{orange!15}22.5 & \cellcolor{red!15}31.4 & \cellcolor{red!15}24.3 \\
\midrule
OpenCLIP XLMR B-32 & No training & 24.4 & 3.1 & 0.7 & 25.8 & 5.8 & 25.8 & 37.4 & 29.8 & 45.1 & 35.2 & 17.1 & 24.6 & 33.8 & 32.7 & 47.8 & 40.9 \\
 \cmidrule(lr){2-2}
 & Text Contrastive & \cellcolor{teal!15}44.0 & \cellcolor{blue!15}26.3 & \cellcolor{teal!15}16.0 & \cellcolor{teal!15}38.7 & \cellcolor{blue!15}26.8 & \cellcolor{cyan!15}32.0 & 37.8 & 30.0 & \cellcolor{red!15}39.0 & \cellcolor{orange!15}31.0 & \cellcolor{cyan!15}26.8 & \cellcolor{orange!15}21.0 & \cellcolor{red!15}25.0 & \cellcolor{red!15}26.6 & \cellcolor{red!15}39.0 & \cellcolor{red!15}33.8 \\
 & LiT & \cellcolor{blue!15}47.7 & \cellcolor{blue!15}33.7 & \cellcolor{teal!15}19.7 & \cellcolor{teal!15}37.5 & \cellcolor{teal!15}23.0 & 30.6 & \cellcolor{orange!15}35.5 & \cellcolor{orange!15}28.2 & \cellcolor{red!15}38.7 & \cellcolor{orange!15}30.4 & \cellcolor{teal!15}28.9 & \cellcolor{orange!15}22.7 & \cellcolor{red!15}27.4 & \cellcolor{red!15}27.7 & \cellcolor{red!15}39.5 & \cellcolor{red!15}30.7 \\
 & MSE & \cellcolor{blue!15}47.6 & \cellcolor{blue!15}38.7 & \cellcolor{blue!15}24.7 & \cellcolor{teal!15}42.8 & \cellcolor{blue!15}30.4 & \cellcolor{teal!15}39.0 & \cellcolor{cyan!15}44.0 & 34.6 & \cellcolor{orange!15}44.2 & 36.7 & \cellcolor{teal!15}36.2 & 26.9 & \cellcolor{orange!15}31.9 & 33.0 & \cellcolor{orange!15}45.9 & \cellcolor{red!15}33.6 \\
% \cmidrule(lr){2-2}
% & MSE (WikiData) & --- & \cellcolor{teal!15}17.1 & --- & --- & --- & \cellcolor{red!15}18.1 & --- & \cellcolor{red!15}18.5 & \cellcolor{red!15}25.0 & \cellcolor{red!15}23.0 & 17.7 & \cellcolor{orange!15}21.1 & \cellcolor{red!15}14.7 & \cellcolor{red!15}23.0 & \cellcolor{red!15}38.6 & \cellcolor{red!15}28.5 \\ \\
\midrule
 \# Classes & & 35  & 97  & 141  & 220  & 232  & 342  & 419  & 496  & 509  & 535  & 625  & 636  & 648  & 682  & 738  & 885 \\
 \# XLM-R Tokens & & 1.11  & 2.39  & 1.23  & 2.86  & 1.85  & 3.23  & 3.12  & 2.93  & 3.55  & 3.26  & 2.43  & 3.46  & 3.75  & 4.12  & 4.01  & 2.64 \\
 \# LAION5B Examples & & 6.71  & 4.11  & 4.07  & 6.13  & 4.49  & 7.18  & 7.05  & 7.01  & 7.06  & 6.98  & 6.73  & 7.35  & 7.01  & 7.32  & 8.18  & 8.16 \\
 M-CLIP Distilled & & T  & F  & F  & T  & F  & T  & F  & T  & F  & F  & F  & T  & F  & F  & T  & T \\
 \bottomrule
    \end{tabular}
    }
    \caption{Results of adapter-based language adaptation of M-CLIP and OpenCLIP with three objectives (Text Contrastive, Text MSE, and LiT). Comparison against the model before adaptation.
    %and (ii) best-performing of all 8 CLIP models (see Table \ref{tbl:models}) for each language. 
    Colors denote the size of change in performance w.r.t. original model: {\color{red}$\leq-5$}, {\color{orange}$\leq0$}, $\leq5$, {\color{cyan}$\leq10$}, {\color{teal}$\leq20$}, {\color{blue}$>20$} (best viewed in color).
    We additionally report language statistics: the number of classes, the number of tokens in XLM-R pre-training (in millions, log10), the number of examples in LAION5B (log10) and whether the language was used in M-CLIP training (True/False).
    }
    \label{tab:res:train_onelang}
\end{table*}

Finally, we improve the performance for low-resource languages by resorting to parameter-efficient fine-tuning with adapters \citep{houlsby_parameter-efficient_2019,pfeiffer_mad-x_2020}, trainable bottleneck layers that we insert into the text encoder. We only update adapter parameters, keeping the original CLIP parameters frozen.
We train a separate adapter for each language. 
%on top of the same CLIP model
% , which is computationally much cheaper than fine-tuning a specialized CLIP model for each language.  

\paragraph{Setup.} We train language-specific adapters on top of (a) OpenClip B-32 model (trained from scratch) and (b) M-CLIP XLMR-L B-32 (obtained via distillation). We experiment with three training objectives: English-target language distillation with (i) MSE and (ii) contrastive loss, and (iii) LiT on image-caption pairs. The former two require parallel data, whereas the latter requires images paired with target-language captions.
For comparability between languages, we 
%follow \citet{carlsson_cross-lingual_2022} and
sample 100K captions (with corresponding images) from the synthetic dataset provided by \citet{li_blip_2022} and translate them automatically to all target languages with NLLB.
%Additionally, we use parallel Wikipedia sentences from WikiMatrix \citep{schwenk_wikimatrix_2021} where available as alternative data source.
We perform adapter-based specialization for 16 languages. One run (i.e., one model-language-objective combination) takes 3h on a single Nvidia RTX 3090 card (see \S\ref{sec:appendix:hyperparameters} for details).

\paragraph{Results.} Table \ref{tab:res:train_onelang} displays the results.
%First, models trained with the WikiMatrix data see a performance decrease throughout (except for \textit{si} with OpenCLIP).
%This suggests that using `non-visual' sentences is not helpful at least in our setup with already-trained models. 
For low-resource languages (\textit{xh}, \textit{si}, \textit{lo}, \textit{my}, and \textit{eu}), we observe massive improvements.
%,  outperforming prior best results by wide margins.
In contrast, the adaptation brings performance losses for high-resource languages (e.g., \textit{de} and \textit{zh}). We hypothesize that constraining the representation space of a target language to English representations is beneficial for low-resource languages with semantically poor initial representations, but detrimental for high-resource languages with semantically accurate initial representations.  
%of under-resource languages to the English representation space (drastically) improves their semantic quality; for high-resource this alignment reduces the quality of already good language-specific representations by (overly) constrain languages for which initial CLIP representations are of high-quality this alignment overly constraints the high-quality target language representation to English. 
%is harmful.
%For mid-resource languages, results are mixed whether we see improvements or not.
%%
%%
%% GG: too hand-wavy
%There is no clear `cut-off' point for XLM-R tokens or LAION5B examples after which results degrade likely due to language-inherent factors (relations to other languages, scripts, etc.).
%At least for M-CLIP, it seems that not-distilled languages are more likely to improve with training (e.g. \textit{ko, fa, lt}) but this is also not always the case (\textit{ms, sk}).
For both OpenCLIP and M-CLIP, adaptation with the MSE objective on parallel sentences yields the best results. 
Overall, the trends in performance changes from language adaptation are very similar between OpenCLIP and M-CLIP, despite the fact that they were obtained using very different training procedures and trained on datasets with different language distributions. This suggests that this commonality in language adaptation behavior stems from the initialization of the text encoder with XLM-R weights.

\section{Conclusion}

We introduced Babel-ImageNet, the first massively multilingual translation of the ImageNet classes to 100 languages.
We leverage the WordNet synsets as the link between ImageNet and BabelNet to obtain high-quality translations without relying on MT or human annotators. 
%%%
Using \bin{}, we carried out the most comprehensively multilingual comparative evaluation of 11 public CLIP models on zero-shot image classification, demonstrating that all models fail for low(er)-resource languages. Crucially, we validate our benchmark by showing that models' text-to-image retrieval performance (on three datasets) strongly correlates with their ZS-IC performance on \bin{} for the corresponding languages.     
%benchmark---2 trained from scratch on LAION5B and the others derived with teacher distillation from English CLIP models---and show that no training approach achives satisfactory results over all languages.
Finally, we proposed a parameter-efficient fine-tuning procedure that drastically improves the performance of multilingual CLIP models for low-resource languages. 
%Motivated by those short-comings, we investigate to what extend we can improve results for one language with limited training, testing different losses and data sources in the process.
%While we achieve significant improvements for low-resource languages, for mid-to-high-resource languages, results are mixed and we can get even worse results after training.

%In our analysis, we investigate the impact of the partial translation on the results. 
%While accuracy changes in a log-linear relationship with the number of classes, using a fixed number of classes does not change the qualitative results.
%We also compare our benchmark results with performance on existing image-text retrieval datasets and show that results on both are highly correlated.

The wide range of languages encompassed by our benchmark reveals that the theoretical ``multilinguality'' of CLIP models is practically very limited and points to the need for methods that derive robust VL encoders with much stronger performance especially for low-resource languages: e.g., better distillation procedures that retain more of the impressive performance of English CLIP. 
%and improved training setups that yield better cross-lingual transfer so that lower-resource languages can close the gap to other languages.

% \newpage

\section{Limitations}
\label{sec:appendix:limitations}
While \bin{}  greatly improves language coverage for evaluation of multilingual VL models, there are some limitations of our work:

For one, ImageNet classes tend to be Anglo-centric due to inherited biases from WordNet \citep{shankar_no_2017,devries_does_2019,liu_visually_2021} so while our benchmark evaluates the performance on languages from all over the globe, we do not evaluate the model performance on concepts specific (or even unique) to cultures in which those languages are spoken.
As a result, \bin{} may overestimate the actual usability of an VL model for real-world uses in some cultures and geographies.

Further, we select for \bin{} the 92 languages used in XLM-R pretraining along with 8 more manually chosen languages. This selection reinforces research focus on those languages to the detriment of other (mainly extremely low-resource) languages. However, we release our code, as well as data for labels of 298 languages and encourage future research to consider an even wider set of languages.

\section*{Acknowledgements}

This work was in part supported by the Alexander von Humboldt Foundation.

\bibliography{anthology,custom}

\begin{thebibliography}{72}
\expandafter\ifx\csname natexlab\endcsname\relax\def\natexlab#1{#1}\fi

\bibitem[{Aggarwal and Kale(2020)}]{aggarwal_towards_2020}
Pranav Aggarwal and Ajinkya Kale. 2020.
\newblock \href {https://arxiv.org/abs/2012.05107} {Towards {Zero}-shot
  {Cross}-lingual {Image} {Retrieval}}.
\newblock \emph{CoRR}, abs/2012.05107.
\newblock ArXiv: 2012.05107.

\bibitem[{Barrault et~al.(2018)Barrault, Bougares, Specia, Lala, Elliott, and
  Frank}]{barrault_findings_2018}
Loïc Barrault, Fethi Bougares, Lucia Specia, Chiraag Lala, Desmond Elliott,
  and Stella Frank. 2018.
\newblock \href {https://doi.org/10.18653/v1/W18-6402} {Findings of the {Third}
  {Shared} {Task} on {Multimodal} {Machine} {Translation}}.
\newblock In \emph{Proceedings of the {Third} {Conference} on {Machine}
  {Translation}: {Shared} {Task} {Papers}}, pages 304--323, Belgium, Brussels.
  Association for Computational Linguistics.

\bibitem[{Bianchi et~al.(2021)Bianchi, Attanasio, Pisoni, Terragni, Sarti, and
  Lakshmi}]{bianchi_contrastive_2021}
Federico Bianchi, Giuseppe Attanasio, Raphael Pisoni, Silvia Terragni, Gabriele
  Sarti, and Sri Lakshmi. 2021.
\newblock \href {https://arxiv.org/abs/2108.08688} {Contrastive
  {Language}-{Image} {Pre}-training for the {Italian} {Language}}.
\newblock \emph{CoRR}, abs/2108.08688.
\newblock ArXiv: 2108.08688.

\bibitem[{Bugliarello et~al.(2022)Bugliarello, Liu, Pfeiffer, Reddy, Elliott,
  Ponti, and Vulic}]{bugliarello_iglue_2022}
Emanuele Bugliarello, Fangyu Liu, Jonas Pfeiffer, Siva Reddy, Desmond Elliott,
  Edoardo~Maria Ponti, and Ivan Vulic. 2022.
\newblock \href {https://proceedings.mlr.press/v162/bugliarello22a.html}
  {{IGLUE:} {A} benchmark for transfer learning across modalities, tasks, and
  languages}.
\newblock In \emph{International Conference on Machine Learning, {ICML} 2022,
  17-23 July 2022, Baltimore, Maryland, {USA}}, volume 162 of \emph{Proceedings
  of Machine Learning Research}, pages 2370--2392. {PMLR}.

\bibitem[{Burns et~al.(2020)Burns, Kim, Wijaya, Saenko, and
  Plummer}]{burns_learning_2020}
Andrea Burns, Donghyun Kim, Derry Wijaya, Kate Saenko, and Bryan~A. Plummer.
  2020.
\newblock \href {https://doi.org/10.1007/978-3-030-58548-8_12} {Learning to
  {Scale} {Multilingual} {Representations} for {Vision}-{Language} {Tasks}}.
\newblock In \emph{Computer {Vision} - {ECCV} 2020 - 16th {European}
  {Conference}, {Glasgow}, {UK}, {August} 23-28, 2020, {Proceedings}, {Part}
  {IV}}, volume 12349 of \emph{Lecture {Notes} in {Computer} {Science}}, pages
  197--213. Springer.

\bibitem[{Carlsson et~al.(2022)Carlsson, Eisen, Rekathati, and
  Sahlgren}]{carlsson_cross-lingual_2022}
Fredrik Carlsson, Philipp Eisen, Faton Rekathati, and Magnus Sahlgren. 2022.
\newblock \href {https://aclanthology.org/2022.lrec-1.739} {Cross-lingual and
  {Multilingual} {CLIP}}.
\newblock In \emph{Proceedings of the {Thirteenth} {Language} {Resources} and
  {Evaluation} {Conference}, {LREC} 2022, {Marseille}, {France}, 20-25 {June}
  2022}, pages 6848--6854. European Language Resources Association.

\bibitem[{Chen et~al.(2022{\natexlab{a}})Chen, Wang, Changpinyo, Piergiovanni,
  Padlewski, Salz, Goodman, Grycner, Mustafa, Beyer, Kolesnikov, Puigcerver,
  Ding, Rong, Akbari, Mishra, Xue, Thapliyal, Bradbury, Kuo, Seyedhosseini,
  Jia, Ayan, Riquelme, Steiner, Angelova, Zhai, Houlsby, and
  Soricut}]{chen_pali_2022}
Xi~Chen, Xiao Wang, Soravit Changpinyo, A.~J. Piergiovanni, Piotr Padlewski,
  Daniel Salz, Sebastian Goodman, Adam Grycner, Basil Mustafa, Lucas Beyer,
  Alexander Kolesnikov, Joan Puigcerver, Nan Ding, Keran Rong, Hassan Akbari,
  Gaurav Mishra, Linting Xue, Ashish Thapliyal, James Bradbury, Weicheng Kuo,
  Mojtaba Seyedhosseini, Chao Jia, Burcu~Karagol Ayan, Carlos Riquelme, Andreas
  Steiner, Anelia Angelova, Xiaohua Zhai, Neil Houlsby, and Radu Soricut.
  2022{\natexlab{a}}.
\newblock \href {https://doi.org/10.48550/arXiv.2209.06794} {{PaLI}: {A}
  {Jointly}-{Scaled} {Multilingual} {Language}-{Image} {Model}}.
\newblock \emph{CoRR}, abs/2209.06794.
\newblock ArXiv: 2209.06794.

\bibitem[{Chen et~al.(2022{\natexlab{b}})Chen, Liu, Zhang, Ye, Yang, and
  Wu}]{chen_altclip_2022}
Zhongzhi Chen, Guang Liu, Bo-Wen Zhang, Fulong Ye, Qinghong Yang, and Ledell
  Wu. 2022{\natexlab{b}}.
\newblock \href {https://doi.org/10.48550/arXiv.2211.06679} {{AltCLIP}:
  {Altering} the {Language} {Encoder} in {CLIP} for {Extended} {Language}
  {Capabilities}}.
\newblock \emph{CoRR}, abs/2211.06679.
\newblock ArXiv: 2211.06679.

\bibitem[{Conneau et~al.(2020)Conneau, Khandelwal, Goyal, Chaudhary, Wenzek,
  Guzmán, Grave, Ott, Zettlemoyer, and Stoyanov}]{conneau_unsupervised_2020}
Alexis Conneau, Kartikay Khandelwal, Naman Goyal, Vishrav Chaudhary, Guillaume
  Wenzek, Francisco Guzmán, Edouard Grave, Myle Ott, Luke Zettlemoyer, and
  Veselin Stoyanov. 2020.
\newblock \href {https://doi.org/10.18653/v1/2020.acl-main.747} {Unsupervised
  {Cross}-lingual {Representation} {Learning} at {Scale}}.
\newblock In \emph{Proceedings of the 58th {Annual} {Meeting} of the
  {Association} for {Computational} {Linguistics}, {ACL} 2020, {Online}, {July}
  5-10, 2020}, pages 8440--8451. Association for Computational Linguistics.

\bibitem[{Costa-jussà et~al.(2022)Costa-jussà, Cross, Çelebi, Elbayad,
  Heafield, Heffernan, Kalbassi, Lam, Licht, Maillard, Sun, Wang, Wenzek,
  Youngblood, Akula, Barrault, Gonzalez, Hansanti, Hoffman, Jarrett, Sadagopan,
  Rowe, Spruit, Tran, Andrews, Ayan, Bhosale, Edunov, Fan, Gao, Goswami,
  Guzmán, Koehn, Mourachko, Ropers, Saleem, Schwenk, and
  Wang}]{costa-jussa_no_2022}
Marta~R. Costa-jussà, James Cross, Onur Çelebi, Maha Elbayad, Kenneth
  Heafield, Kevin Heffernan, Elahe Kalbassi, Janice Lam, Daniel Licht, Jean
  Maillard, Anna Sun, Skyler Wang, Guillaume Wenzek, Al~Youngblood, Bapi Akula,
  Loïc Barrault, Gabriel~Mejia Gonzalez, Prangthip Hansanti, John Hoffman,
  Semarley Jarrett, Kaushik~Ram Sadagopan, Dirk Rowe, Shannon Spruit, Chau
  Tran, Pierre Andrews, Necip~Fazil Ayan, Shruti Bhosale, Sergey Edunov, Angela
  Fan, Cynthia Gao, Vedanuj Goswami, Francisco Guzmán, Philipp Koehn,
  Alexandre Mourachko, Christophe Ropers, Safiyyah Saleem, Holger Schwenk, and
  Jeff Wang. 2022.
\newblock \href {https://doi.org/10.48550/arXiv.2207.04672} {No {Language}
  {Left} {Behind}: {Scaling} {Human}-{Centered} {Machine} {Translation}}.
\newblock \emph{CoRR}, abs/2207.04672.
\newblock ArXiv: 2207.04672.

\bibitem[{Deng et~al.(2009)Deng, Dong, Socher, Li, {Kai Li}, and {Li
  Fei-Fei}}]{deng_imagenet_2009}
J.~Deng, W.~Dong, R.~Socher, L.~Li, {Kai Li}, and {Li Fei-Fei}. 2009.
\newblock \href {https://doi.org/10.1109/CVPR.2009.5206848} {{ImageNet}: {A}
  large-scale hierarchical image database}.
\newblock In \emph{2009 {IEEE} {Conference} on {Computer} {Vision} and
  {Pattern} {Recognition}}, pages 248--255.

\bibitem[{Devlin et~al.(2019)Devlin, Chang, Lee, and
  Toutanova}]{devlin_bert_2019}
Jacob Devlin, Ming-Wei Chang, Kenton Lee, and Kristina Toutanova. 2019.
\newblock \href {https://doi.org/10.18653/v1/n19-1423} {{BERT}: {Pre}-training
  of {Deep} {Bidirectional} {Transformers} for {Language} {Understanding}}.
\newblock In \emph{Proceedings of the 2019 {Conference} of the {North}
  {American} {Chapter} of the {Association} for {Computational} {Linguistics}:
  {Human} {Language} {Technologies}, {NAACL}-{HLT} 2019, {Minneapolis}, {MN},
  {USA}, {June} 2-7, 2019, {Volume} 1 ({Long} and {Short} {Papers})}, pages
  4171--4186. Association for Computational Linguistics.

\bibitem[{DeVries et~al.(2019)DeVries, Misra, Wang, and
  Maaten}]{devries_does_2019}
Terrance DeVries, Ishan Misra, Changhan Wang, and Laurens van~der Maaten. 2019.
\newblock \href
  {http://openaccess.thecvf.com/content\_CVPRW\_2019/html/cv4gc/de\_Vries\_Does\_Object\_Recognition\_Work\_for\_Everyone\_CVPRW\_2019\_paper.html}
  {Does {Object} {Recognition} {Work} for {Everyone}?}
\newblock In \emph{{IEEE} {Conference} on {Computer} {Vision} and {Pattern}
  {Recognition} {Workshops}, {CVPR} {Workshops} 2019, {Long} {Beach}, {CA},
  {USA}, {June} 16-20, 2019}, pages 52--59. Computer Vision Foundation / IEEE.

\bibitem[{Dosovitskiy et~al.(2021)Dosovitskiy, Beyer, Kolesnikov, Weissenborn,
  Zhai, Unterthiner, Dehghani, Minderer, Heigold, Gelly, Uszkoreit, and
  Houlsby}]{dosovitskiy_image_2020}
Alexey Dosovitskiy, Lucas Beyer, Alexander Kolesnikov, Dirk Weissenborn,
  Xiaohua Zhai, Thomas Unterthiner, Mostafa Dehghani, Matthias Minderer, Georg
  Heigold, Sylvain Gelly, Jakob Uszkoreit, and Neil Houlsby. 2021.
\newblock \href {https://openreview.net/forum?id=YicbFdNTTy} {An image is worth
  16x16 words: Transformers for image recognition at scale}.
\newblock In \emph{9th International Conference on Learning Representations,
  {ICLR} 2021, Virtual Event, Austria, May 3-7, 2021}. OpenReview.net.

\bibitem[{Eichenberg et~al.(2022)Eichenberg, Black, Weinbach, Parcalabescu, and
  Frank}]{eichenberg_magma_2021}
Constantin Eichenberg, Sidney Black, Samuel Weinbach, Letitia Parcalabescu, and
  Anette Frank. 2022.
\newblock \href {https://aclanthology.org/2022.findings-emnlp.179} {{MAGMA} -
  multimodal augmentation of generative models through adapter-based
  finetuning}.
\newblock In \emph{Findings of the Association for Computational Linguistics:
  {EMNLP} 2022, Abu Dhabi, United Arab Emirates, December 7-11, 2022}, pages
  2416--2428. Association for Computational Linguistics.

\bibitem[{Elliott et~al.(2017)Elliott, Frank, Barrault, Bougares, and
  Specia}]{elliott_findings_2017}
Desmond Elliott, Stella Frank, Loïc Barrault, Fethi Bougares, and Lucia
  Specia. 2017.
\newblock \href {https://doi.org/10.18653/v1/W17-4718} {Findings of the
  {Second} {Shared} {Task} on {Multimodal} {Machine} {Translation} and
  {Multilingual} {Image} {Description}}.
\newblock In \emph{Proceedings of the {Second} {Conference} on {Machine}
  {Translation}}, pages 215--233, Copenhagen, Denmark. Association for
  Computational Linguistics.

\bibitem[{Elliott et~al.(2016)Elliott, Frank, Sima'an, and
  Specia}]{elliott_multi30k_2016}
Desmond Elliott, Stella Frank, Khalil Sima'an, and Lucia Specia. 2016.
\newblock \href {https://doi.org/10.18653/v1/W16-3210} {{Multi30K}:
  {Multilingual} {English}-{German} {Image} {Descriptions}}.
\newblock In \emph{Proceedings of the 5th {Workshop} on {Vision} and
  {Language}}, pages 70--74, Berlin, Germany. Association for Computational
  Linguistics.

\bibitem[{Faghri et~al.(2018)Faghri, Fleet, Kiros, and
  Fidler}]{faghri_vse_2018}
Fartash Faghri, David~J. Fleet, Jamie~Ryan Kiros, and Sanja Fidler. 2018.
\newblock \href {http://bmvc2018.org/contents/papers/0344.pdf} {{VSE}++:
  {Improving} {Visual}-{Semantic} {Embeddings} with {Hard} {Negatives}}.
\newblock In \emph{British {Machine} {Vision} {Conference} 2018, {BMVC} 2018,
  {Newcastle}, {UK}, {September} 3-6, 2018}, page~12. BMVA Press.

\bibitem[{Fang et~al.(2023)Fang, Kornblith, and Schmidt}]{fang_does_2023}
Alex Fang, Simon Kornblith, and Ludwig Schmidt. 2023.
\newblock \href {https://doi.org/10.48550/arXiv.2301.04644} {Does progress on
  {ImageNet} transfer to real-world datasets?}
\newblock \emph{CoRR}, abs/2301.04644.
\newblock ArXiv: 2301.04644.

\bibitem[{Gao et~al.(2021)Gao, Yao, and Chen}]{gao_simcse_2021}
Tianyu Gao, Xingcheng Yao, and Danqi Chen. 2021.
\newblock \href {https://doi.org/10.18653/v1/2021.emnlp-main.552} {{SimCSE}:
  {Simple} {Contrastive} {Learning} of {Sentence} {Embeddings}}.
\newblock In \emph{Proceedings of the 2021 {Conference} on {Empirical}
  {Methods} in {Natural} {Language} {Processing}, {EMNLP} 2021, {Virtual}
  {Event} / {Punta} {Cana}, {Dominican} {Republic}, 7-11 {November}, 2021},
  pages 6894--6910. Association for Computational Linguistics.

\bibitem[{Geigle et~al.(2022)Geigle, Pfeiffer, Reimers, Vulic, and
  Gurevych}]{geigle_retrieve_2022}
Gregor Geigle, Jonas Pfeiffer, Nils Reimers, Ivan Vulic, and Iryna Gurevych.
  2022.
\newblock \href {https://doi.org/10.1162/tacl_a_00473} {Retrieve {Fast},
  {Rerank} {Smart}: {Cooperative} and {Joint} {Approaches} for {Improved}
  {Cross}-{Modal} {Retrieval}}.
\newblock \emph{Transactions of the Association for Computational Linguistics},
  10:503--521.

\bibitem[{Gella et~al.(2017)Gella, Sennrich, Keller, and
  Lapata}]{gella_image_2017}
Spandana Gella, Rico Sennrich, Frank Keller, and Mirella Lapata. 2017.
\newblock \href {https://doi.org/10.18653/v1/d17-1303} {Image {Pivoting} for
  {Learning} {Multilingual} {Multimodal} {Representations}}.
\newblock In \emph{Proceedings of the 2017 {Conference} on {Empirical}
  {Methods} in {Natural} {Language} {Processing}, {EMNLP} 2017, {Copenhagen},
  {Denmark}, {September} 9-11, 2017}, pages 2839--2845. Association for
  Computational Linguistics.

\bibitem[{Hamp and Feldweg(1997)}]{hamp1997germanet}
Birgit Hamp and Helmut Feldweg. 1997.
\newblock Germanet-a lexical-semantic net for german.
\newblock In \emph{Automatic information extraction and building of lexical
  semantic resources for NLP applications}.

\bibitem[{Houlsby et~al.(2019)Houlsby, Giurgiu, Jastrzebski, Morrone,
  Laroussilhe, Gesmundo, Attariyan, and
  Gelly}]{houlsby_parameter-efficient_2019}
Neil Houlsby, Andrei Giurgiu, Stanislaw Jastrzebski, Bruna Morrone, Quentin~de
  Laroussilhe, Andrea Gesmundo, Mona Attariyan, and Sylvain Gelly. 2019.
\newblock \href {http://proceedings.mlr.press/v97/houlsby19a.html}
  {Parameter-{Efficient} {Transfer} {Learning} for {NLP}}.
\newblock In \emph{Proceedings of the 36th {International} {Conference} on
  {Machine} {Learning}, {ICML} 2019, 9-15 {June} 2019, {Long} {Beach},
  {California}, {USA}}, volume~97 of \emph{Proceedings of {Machine} {Learning}
  {Research}}, pages 2790--2799. PMLR.

\bibitem[{Hu et~al.(2022)Hu, Shen, Wallis, Allen-Zhu, Li, Wang, Wang, and
  Chen}]{hu_lora_2022}
Edward~J. Hu, Yelong Shen, Phillip Wallis, Zeyuan Allen-Zhu, Yuanzhi Li, Shean
  Wang, Lu~Wang, and Weizhu Chen. 2022.
\newblock \href {https://openreview.net/forum?id=nZeVKeeFYf9} {{LoRA}:
  {Low}-{Rank} {Adaptation} of {Large} {Language} {Models}}.
\newblock In \emph{The {Tenth} {International} {Conference} on {Learning}
  {Representations}, {ICLR} 2022, {Virtual} {Event}, {April} 25-29, 2022}.
  OpenReview.net.

\bibitem[{Hu et~al.(2020)Hu, Ruder, Siddhant, Neubig, Firat, and
  Johnson}]{hu2020xtreme}
Junjie Hu, Sebastian Ruder, Aditya Siddhant, Graham Neubig, Orhan Firat, and
  Melvin Johnson. 2020.
\newblock Xtreme: A massively multilingual multi-task benchmark for evaluating
  cross-lingual generalisation.
\newblock In \emph{International Conference on Machine Learning}, pages
  4411--4421. PMLR.

\bibitem[{Ilharco et~al.(2021)Ilharco, Wortsman, Wightman, Gordon, Carlini,
  Taori, Dave, Shankar, Namkoong, Miller, Hajishirzi, Farhadi, and
  Schmidt}]{ilharco_openclip_2021}
Gabriel Ilharco, Mitchell Wortsman, Ross Wightman, Cade Gordon, Nicholas
  Carlini, Rohan Taori, Achal Dave, Vaishaal Shankar, Hongseok Namkoong, John
  Miller, Hannaneh Hajishirzi, Ali Farhadi, and Ludwig Schmidt. 2021.
\newblock \href {https://doi.org/10.5281/zenodo.5143773} {{OpenCLIP}}.

\bibitem[{Jain et~al.(2021)Jain, Guo, Srinivasan, Chen, Kudugunta, Jia, Yang,
  and Baldridge}]{jain_mural_2021}
Aashi Jain, Mandy Guo, Krishna Srinivasan, Ting Chen, Sneha Kudugunta, Chao
  Jia, Yinfei Yang, and Jason Baldridge. 2021.
\newblock \href {https://doi.org/10.18653/v1/2021.findings-emnlp.293} {{MURAL}:
  {Multimodal}, {Multitask} {Representations} {Across} {Languages}}.
\newblock In \emph{Findings of the {Association} for {Computational}
  {Linguistics}: {EMNLP} 2021, {Virtual} {Event} / {Punta} {Cana}, {Dominican}
  {Republic}, 16-20 {November}, 2021}, pages 3449--3463. Association for
  Computational Linguistics.

\bibitem[{Jia et~al.(2021)Jia, Yang, Xia, Chen, Parekh, Pham, Le, Sung, Li, and
  Duerig}]{jia_scaling_2021}
Chao Jia, Yinfei Yang, Ye~Xia, Yi-Ting Chen, Zarana Parekh, Hieu Pham, Quoc~V.
  Le, Yun-Hsuan Sung, Zhen Li, and Tom Duerig. 2021.
\newblock \href {http://proceedings.mlr.press/v139/jia21b.html} {Scaling {Up}
  {Visual} and {Vision}-{Language} {Representation} {Learning} {With} {Noisy}
  {Text} {Supervision}}.
\newblock In \emph{Proceedings of the 38th {International} {Conference} on
  {Machine} {Learning}, {ICML} 2021, 18-24 {July} 2021, {Virtual} {Event}},
  volume 139 of \emph{Proceedings of {Machine} {Learning} {Research}}, pages
  4904--4916. PMLR.

\bibitem[{Karpathy and Fei-Fei(2017)}]{karpathy_deep_2017}
A.~Karpathy and L.~Fei-Fei. 2017.
\newblock \href {https://doi.org/10.1109/TPAMI.2016.2598339} {Deep
  {Visual}-{Semantic} {Alignments} for {Generating} {Image} {Descriptions}}.
\newblock \emph{IEEE Transactions on Pattern Analysis and Machine
  Intelligence}, 39(4):664--676.

\bibitem[{Kim et~al.(2020)Kim, Saito, Saenko, Sclaroff, and
  Plummer}]{kim_mule_2020}
Donghyun Kim, Kuniaki Saito, Kate Saenko, Stan Sclaroff, and Bryan~A. Plummer.
  2020.
\newblock \href {https://aaai.org/ojs/index.php/AAAI/article/view/6785}
  {{MULE}: {Multimodal} {Universal} {Language} {Embedding}}.
\newblock In \emph{The {Thirty}-{Fourth} {AAAI} {Conference} on {Artificial}
  {Intelligence}, {AAAI} 2020, {The} {Thirty}-{Second} {Innovative}
  {Applications} of {Artificial} {Intelligence} {Conference}, {IAAI} 2020,
  {The} {Tenth} {AAAI} {Symposium} on {Educational} {Advances} in {Artificial}
  {Intelligence}, {EAAI} 2020, {New} {York}, {NY}, {USA}, {February} 7-12,
  2020}, pages 11254--11261. AAAI Press.

\bibitem[{Lauscher et~al.(2020)Lauscher, Ravishankar, Vuli{\'c}, and
  Glava{\v{s}}}]{lauscher2020zero}
Anne Lauscher, Vinit Ravishankar, Ivan Vuli{\'c}, and Goran Glava{\v{s}}. 2020.
\newblock From zero to hero: On the limitations of zero-shot language transfer
  with multilingual transformers.
\newblock In \emph{Proceedings of the 2020 Conference on Empirical Methods in
  Natural Language Processing (EMNLP)}, pages 4483--4499.

\bibitem[{Li et~al.(2023)Li, Li, Savarese, and Hoi}]{li_blip-2_2023}
Junnan Li, Dongxu Li, Silvio Savarese, and Steven C.~H. Hoi. 2023.
\newblock \href {https://doi.org/10.48550/arXiv.2301.12597} {{BLIP}-2:
  {Bootstrapping} {Language}-{Image} {Pre}-training with {Frozen} {Image}
  {Encoders} and {Large} {Language} {Models}}.
\newblock \emph{CoRR}, abs/2301.12597.
\newblock ArXiv: 2301.12597.

\bibitem[{Li et~al.(2022)Li, Li, Xiong, and Hoi}]{li_blip_2022}
Junnan Li, Dongxu Li, Caiming Xiong, and Steven C.~H. Hoi. 2022.
\newblock \href {https://proceedings.mlr.press/v162/li22n.html} {{BLIP}:
  {Bootstrapping} {Language}-{Image} {Pre}-training for {Unified}
  {Vision}-{Language} {Understanding} and {Generation}}.
\newblock In \emph{International {Conference} on {Machine} {Learning}, {ICML}
  2022, 17-23 {July} 2022, {Baltimore}, {Maryland}, {USA}}, volume 162 of
  \emph{Proceedings of {Machine} {Learning} {Research}}, pages 12888--12900.
  PMLR.

\bibitem[{Li et~al.(2019)Li, Xu, Wang, Lan, Jia, Yang, and
  Xu}]{li_coco-cn_2019}
Xirong Li, Chaoxi Xu, Xiaoxu Wang, Weiyu Lan, Zhengxiong Jia, Gang Yang, and
  Jieping Xu. 2019.
\newblock \href {https://doi.org/10.1109/TMM.2019.2896494} {{COCO}-{CN} for
  {Cross}-{Lingual} {Image} {Tagging}, {Captioning}, and {Retrieval}}.
\newblock \emph{IEEE Trans. Multim.}, 21(9):2347--2360.

\bibitem[{Lin et~al.(2014)Lin, Maire, Belongie, Hays, Perona, Ramanan, Dollár,
  and Zitnick}]{lin_microsoft_2014}
Tsung-Yi Lin, Michael Maire, Serge~J. Belongie, James Hays, Pietro Perona, Deva
  Ramanan, Piotr Dollár, and C.~Lawrence Zitnick. 2014.
\newblock \href {https://doi.org/10.1007/978-3-319-10602-1_48} {Microsoft
  {COCO}: {Common} {Objects} in {Context}}.
\newblock In \emph{Computer {Vision} - {ECCV} 2014 - 13th {European}
  {Conference}, {Zurich}, {Switzerland}, {September} 6-12, 2014, {Proceedings},
  {Part} {V}}, volume 8693 of \emph{Lecture {Notes} in {Computer} {Science}},
  pages 740--755. Springer.

\bibitem[{Liu et~al.(2021)Liu, Bugliarello, Ponti, Reddy, Collier, and
  Elliott}]{liu_visually_2021}
Fangyu Liu, Emanuele Bugliarello, Edoardo~Maria Ponti, Siva Reddy, Nigel
  Collier, and Desmond Elliott. 2021.
\newblock \href {https://doi.org/10.18653/v1/2021.emnlp-main.818} {Visually
  {Grounded} {Reasoning} across {Languages} and {Cultures}}.
\newblock In \emph{Proceedings of the 2021 {Conference} on {Empirical}
  {Methods} in {Natural} {Language} {Processing}, {EMNLP} 2021, {Virtual}
  {Event} / {Punta} {Cana}, {Dominican} {Republic}, 7-11 {November}, 2021},
  pages 10467--10485. Association for Computational Linguistics.

\bibitem[{Loshchilov and Hutter(2019)}]{loshchilov_decoupled_2019}
Ilya Loshchilov and Frank Hutter. 2019.
\newblock \href {https://openreview.net/forum?id=Bkg6RiCqY7} {Decoupled
  {Weight} {Decay} {Regularization}}.
\newblock In \emph{7th {International} {Conference} on {Learning}
  {Representations}, {ICLR} 2019, {New} {Orleans}, {LA}, {USA}, {May} 6-9,
  2019}. OpenReview.net.

\bibitem[{Miller(1994)}]{miller_wordnet_1994}
George~A. Miller. 1994.
\newblock \href {https://aclanthology.org/H94-1111} {{WordNet}: {A} {Lexical}
  {Database} for {English}}.
\newblock In \emph{Human {Language} {Technology}: {Proceedings} of a {Workshop}
  held at {Plainsboro}, {New} {Jersey}, {March} 8-11, 1994}.

\bibitem[{Navigli et~al.(2021)Navigli, Bevilacqua, Conia, Montagnini, and
  Cecconi}]{navigali_ten_2021}
Roberto Navigli, Michele Bevilacqua, Simone Conia, Dario Montagnini, and
  Francesco Cecconi. 2021.
\newblock \href {https://doi.org/10.24963/IJCAI.2021/620} {Ten years of
  babelnet: {A} survey}.
\newblock In \emph{Proceedings of the Thirtieth International Joint Conference
  on Artificial Intelligence, {IJCAI} 2021, Virtual Event / Montreal, Canada,
  19-27 August 2021}, pages 4559--4567. ijcai.org.

\bibitem[{Navigli and Ponzetto(2010)}]{navigli_babelnet_2010}
Roberto Navigli and Simone~Paolo Ponzetto. 2010.
\newblock \href {https://aclanthology.org/P10-1023/} {{BabelNet}: {Building} a
  {Very} {Large} {Multilingual} {Semantic} {Network}}.
\newblock In \emph{{ACL} 2010, {Proceedings} of the 48th {Annual} {Meeting} of
  the {Association} for {Computational} {Linguistics}, {July} 11-16, 2010,
  {Uppsala}, {Sweden}}, pages 216--225. The Association for Computer
  Linguistics.

\bibitem[{Ni et~al.(2021)Ni, Huang, Su, Cui, Bharti, Wang, Zhang, and
  Duan}]{huang_m3p_2020}
Minheng Ni, Haoyang Huang, Lin Su, Edward Cui, Taroon Bharti, Lijuan Wang,
  Dongdong Zhang, and Nan Duan. 2021.
\newblock \href {https://doi.org/10.1109/CVPR46437.2021.00397} {{M3P}:
  {Learning} {Universal} {Representations} via {Multitask} {Multilingual}
  {Multimodal} {Pre}-{Training}}.
\newblock In \emph{{IEEE} {Conference} on {Computer} {Vision} and {Pattern}
  {Recognition}, {CVPR} 2021, virtual, {June} 19-25, 2021}, pages 3977--3986.
  Computer Vision Foundation / IEEE.

\bibitem[{Nielsen(2018)}]{nielsen_linking_2018}
Finn~{\AA}rup Nielsen. 2018.
\newblock \href {https://doi.org/10.1145/3184558.3191645} {Linking {ImageNet}
  {WordNet} {Synsets} with {Wikidata}}.
\newblock In \emph{Companion of the {The} {Web} {Conference} 2018 on {The}
  {Web} {Conference} 2018, {WWW} 2018, {Lyon} , {France}, {April} 23-27, 2018},
  pages 1809--1814. ACM.

\bibitem[{Nilsback and Zisserman(2008)}]{nilsback_automated_2008}
Maria-Elena Nilsback and Andrew Zisserman. 2008.
\newblock \href {https://doi.org/10.1109/ICVGIP.2008.47} {Automated {Flower}
  {Classification} over a {Large} {Number} of {Classes}}.
\newblock In \emph{Sixth {Indian} {Conference} on {Computer} {Vision},
  {Graphics} \& {Image} {Processing}, {ICVGIP} 2008, {Bhubaneswar}, {India},
  16-19 {December} 2008}, pages 722--729. IEEE Computer Society.

\bibitem[{Northcutt et~al.(2021)Northcutt, Athalye, and
  Mueller}]{northcutt_pervasive_2021}
Curtis~G. Northcutt, Anish Athalye, and Jonas Mueller. 2021.
\newblock \href
  {https://datasets-benchmarks-proceedings.neurips.cc/paper/2021/hash/f2217062e9a397a1dca429e7d70bc6ca-Abstract-round1.html}
  {Pervasive {Label} {Errors} in {Test} {Sets} {Destabilize} {Machine}
  {Learning} {Benchmarks}}.
\newblock In \emph{Proceedings of the {Neural} {Information} {Processing}
  {Systems} {Track} on {Datasets} and {Benchmarks} 1, {NeurIPS} {Datasets} and
  {Benchmarks} 2021, {December} 2021, virtual}.

\bibitem[{Parkhi et~al.(2012)Parkhi, Vedaldi, Zisserman, and
  Jawahar}]{parkhi_cats_2012}
Omkar~M. Parkhi, Andrea Vedaldi, Andrew Zisserman, and C.~V. Jawahar. 2012.
\newblock \href {https://doi.org/10.1109/CVPR.2012.6248092} {Cats and dogs}.
\newblock In \emph{2012 {IEEE} {Conference} on {Computer} {Vision} and
  {Pattern} {Recognition}, {Providence}, {RI}, {USA}, {June} 16-21, 2012},
  pages 3498--3505. IEEE Computer Society.

\bibitem[{Pfeiffer et~al.(2022)Pfeiffer, Geigle, Kamath, Steitz, Roth, Vulic,
  and Gurevych}]{pfeiffer_xgqa_2021}
Jonas Pfeiffer, Gregor Geigle, Aishwarya Kamath, Jan-Martin~O. Steitz, Stefan
  Roth, Ivan Vulic, and Iryna Gurevych. 2022.
\newblock \href {https://doi.org/10.18653/v1/2022.findings-acl.196} {{xGQA}:
  {Cross}-{Lingual} {Visual} {Question} {Answering}}.
\newblock In \emph{Findings of the {Association} for {Computational}
  {Linguistics}: {ACL} 2022, {Dublin}, {Ireland}, {May} 22-27, 2022}, pages
  2497--2511. Association for Computational Linguistics.

\bibitem[{Pfeiffer et~al.(2020{\natexlab{a}})Pfeiffer, Rücklé, Poth, Kamath,
  Vulic, Ruder, Cho, and Gurevych}]{pfeiffer_adapterhub_2020}
Jonas Pfeiffer, Andreas Rücklé, Clifton Poth, Aishwarya Kamath, Ivan Vulic,
  Sebastian Ruder, Kyunghyun Cho, and Iryna Gurevych. 2020{\natexlab{a}}.
\newblock \href {https://doi.org/10.18653/v1/2020.emnlp-demos.7} {{AdapterHub}:
  {A} {Framework} for {Adapting} {Transformers}}.
\newblock In \emph{Proceedings of the 2020 {Conference} on {Empirical}
  {Methods} in {Natural} {Language} {Processing}: {System} {Demonstrations},
  {EMNLP} 2020 - {Demos}, {Online}, {November} 16-20, 2020}, pages 46--54.
  Association for Computational Linguistics.

\bibitem[{Pfeiffer et~al.(2020{\natexlab{b}})Pfeiffer, Vulic, Gurevych, and
  Ruder}]{pfeiffer_mad-x_2020}
Jonas Pfeiffer, Ivan Vulic, Iryna Gurevych, and Sebastian Ruder.
  2020{\natexlab{b}}.
\newblock \href {https://doi.org/10.18653/v1/2020.emnlp-main.617} {{MAD}-{X}:
  {An} {Adapter}-{Based} {Framework} for {Multi}-{Task} {Cross}-{Lingual}
  {Transfer}}.
\newblock In \emph{Proceedings of the 2020 {Conference} on {Empirical}
  {Methods} in {Natural} {Language} {Processing}, {EMNLP} 2020, {Online},
  {November} 16-20, 2020}, pages 7654--7673. Association for Computational
  Linguistics.

\bibitem[{Pham et~al.(2021)Pham, Dai, Ghiasi, Liu, Yu, Luong, Tan, and
  Le}]{pham_combined_2021}
Hieu Pham, Zihang Dai, Golnaz Ghiasi, Hanxiao Liu, Adams~Wei Yu, Minh-Thang
  Luong, Mingxing Tan, and Quoc~V. Le. 2021.
\newblock \href {https://arxiv.org/abs/2111.10050} {Combined {Scaling} for
  {Zero}-shot {Transfer} {Learning}}.
\newblock \emph{CoRR}, abs/2111.10050.
\newblock ArXiv: 2111.10050.

\bibitem[{Pianta et~al.(2002)Pianta, Bentivogli, and
  Girardi}]{pianta2002multiwordnet}
Emanuele Pianta, Luisa Bentivogli, and Christian Girardi. 2002.
\newblock Multiwordnet: developing an aligned multilingual database.
\newblock In \emph{First international conference on global WordNet}, pages
  293--302.

\bibitem[{Plummer et~al.(2015)Plummer, Wang, Cervantes, Caicedo, Hockenmaier,
  and Lazebnik}]{plummer_flickr30k_2015}
Bryan~A. Plummer, Liwei Wang, Chris~M. Cervantes, Juan~C. Caicedo, Julia
  Hockenmaier, and Svetlana Lazebnik. 2015.
\newblock \href {https://doi.org/10.1109/ICCV.2015.303} {Flickr30k {Entities}:
  {Collecting} {Region}-to-{Phrase} {Correspondences} for {Richer}
  {Image}-to-{Sentence} {Models}}.
\newblock In \emph{2015 {IEEE} {International} {Conference} on {Computer}
  {Vision}, {ICCV} 2015, {Santiago}, {Chile}, {December} 7-13, 2015}, pages
  2641--2649.

\bibitem[{Radford et~al.(2021)Radford, Kim, Hallacy, Ramesh, Goh, Agarwal,
  Sastry, Askell, Mishkin, Clark, Krueger, and
  Sutskever}]{radford_learning_2021}
Alec Radford, Jong~Wook Kim, Chris Hallacy, Aditya Ramesh, Gabriel Goh,
  Sandhini Agarwal, Girish Sastry, Amanda Askell, Pamela Mishkin, Jack Clark,
  Gretchen Krueger, and Ilya Sutskever. 2021.
\newblock \href {http://proceedings.mlr.press/v139/radford21a.html} {Learning
  {Transferable} {Visual} {Models} {From} {Natural} {Language} {Supervision}}.
\newblock In \emph{Proceedings of the 38th {International} {Conference} on
  {Machine} {Learning}, {ICML} 2021, 18-24 {July} 2021, {Virtual} {Event}},
  volume 139 of \emph{Proceedings of {Machine} {Learning} {Research}}, pages
  8748--8763. PMLR.

\bibitem[{Recht et~al.(2019)Recht, Roelofs, Schmidt, and
  Shankar}]{recht_imagenet_2019}
Benjamin Recht, Rebecca Roelofs, Ludwig Schmidt, and Vaishaal Shankar. 2019.
\newblock \href {http://proceedings.mlr.press/v97/recht19a.html} {Do {ImageNet}
  {Classifiers} {Generalize} to {ImageNet}?}
\newblock In \emph{Proceedings of the 36th {International} {Conference} on
  {Machine} {Learning}, {ICML} 2019, 9-15 {June} 2019, {Long} {Beach},
  {California}, {USA}}, volume~97 of \emph{Proceedings of {Machine} {Learning}
  {Research}}, pages 5389--5400. PMLR.

\bibitem[{Reimers and Gurevych(2020)}]{reimers_making_2020}
Nils Reimers and Iryna Gurevych. 2020.
\newblock \href {https://doi.org/10.18653/v1/2020.emnlp-main.365} {Making
  {Monolingual} {Sentence} {Embeddings} {Multilingual} using {Knowledge}
  {Distillation}}.
\newblock In \emph{Proceedings of the 2020 {Conference} on {Empirical}
  {Methods} in {Natural} {Language} {Processing}, {EMNLP} 2020, {Online},
  {November} 16-20, 2020}, pages 4512--4525. Association for Computational
  Linguistics.

\bibitem[{Rombach et~al.(2022)Rombach, Blattmann, Lorenz, Esser, and
  Ommer}]{rombach_high-resolution_2022}
Robin Rombach, Andreas Blattmann, Dominik Lorenz, Patrick Esser, and Björn
  Ommer. 2022.
\newblock \href {https://doi.org/10.1109/CVPR52688.2022.01042}
  {High-{Resolution} {Image} {Synthesis} with {Latent} {Diffusion} {Models}}.
\newblock In \emph{{IEEE}/{CVF} {Conference} on {Computer} {Vision} and
  {Pattern} {Recognition}, {CVPR} 2022, {New} {Orleans}, {LA}, {USA}, {June}
  18-24, 2022}, pages 10674--10685. IEEE.

\bibitem[{Schuhmann et~al.(2022)Schuhmann, Beaumont, Vencu, Gordon, Wightman,
  Cherti, Coombes, Katta, Mullis, Wortsman, Schramowski, Kundurthy, Crowson,
  Schmidt, Kaczmarczyk, and Jitsev}]{schuhmann_laion-5b_2022}
Christoph Schuhmann, Romain Beaumont, Richard Vencu, Cade Gordon, Ross
  Wightman, Mehdi Cherti, Theo Coombes, Aarush Katta, Clayton Mullis, Mitchell
  Wortsman, Patrick Schramowski, Srivatsa Kundurthy, Katherine Crowson, Ludwig
  Schmidt, Robert Kaczmarczyk, and Jenia Jitsev. 2022.
\newblock \href
  {http://papers.nips.cc/paper\_files/paper/2022/hash/a1859debfb3b59d094f3504d5ebb6c25-Abstract-Datasets\_and\_Benchmarks.html}
  {{LAION}-{5B}: {An} open large-scale dataset for training next generation
  image-text models}.
\newblock In \emph{{NeurIPS}}.

\bibitem[{Schuhmann et~al.(2021)Schuhmann, Vencu, Beaumont, Kaczmarczyk,
  Mullis, Katta, Coombes, Jitsev, and Komatsuzaki}]{schuhmann_laion-400m_2021}
Christoph Schuhmann, Richard Vencu, Romain Beaumont, Robert Kaczmarczyk,
  Clayton Mullis, Aarush Katta, Theo Coombes, Jenia Jitsev, and Aran
  Komatsuzaki. 2021.
\newblock \href {https://arxiv.org/abs/2111.02114} {{LAION}-{400M}: {Open}
  {Dataset} of {CLIP}-{Filtered} 400 {Million} {Image}-{Text} {Pairs}}.
\newblock \emph{CoRR}, abs/2111.02114.
\newblock ArXiv: 2111.02114.

\bibitem[{Schwenk et~al.(2021)Schwenk, Chaudhary, Sun, Gong, and
  Guzm{\'{a}}n}]{schwenk_wikimatrix_2021}
Holger Schwenk, Vishrav Chaudhary, Shuo Sun, Hongyu Gong, and Francisco
  Guzm{\'{a}}n. 2021.
\newblock \href {https://doi.org/10.18653/v1/2021.eacl-main.115} {Wikimatrix:
  Mining 135m parallel sentences in 1620 language pairs from wikipedia}.
\newblock In \emph{Proceedings of the 16th Conference of the European Chapter
  of the Association for Computational Linguistics: Main Volume, {EACL} 2021,
  Online, April 19 - 23, 2021}, pages 1351--1361. Association for Computational
  Linguistics.

\bibitem[{Shankar et~al.(2017)Shankar, Halpern, Breck, Atwood, Wilson, and
  Sculley}]{shankar_no_2017}
Shreya Shankar, Yoni Halpern, Eric Breck, James Atwood, Jimbo Wilson, and
  D.~Sculley. 2017.
\newblock \href {https://arxiv.org/abs/1711.08536} {No {Classification} without
  {Representation}: {Assessing} {Geodiversity} {Issues} in {Open} {Data} {Sets}
  for the {Developing} {World}}.
\newblock \emph{CoRR}, abs/1711.08536.
\newblock Arxiv: 1711.08536.

\bibitem[{Srinivasan et~al.(2021)Srinivasan, Raman, Chen, Bendersky, and
  Najork}]{srinivasan_wit_2021}
Krishna Srinivasan, Karthik Raman, Jiecao Chen, Michael Bendersky, and Marc
  Najork. 2021.
\newblock \href {https://doi.org/10.1145/3404835.3463257} {{WIT}:
  {Wikipedia}-based {Image} {Text} {Dataset} for {Multimodal} {Multilingual}
  {Machine} {Learning}}.
\newblock In \emph{Proceedings of the 44th {International} {ACM} {SIGIR}
  {Conference} on {Research} and {Development} in {Information} {Retrieval}},
  {SIGIR} '21, pages 2443--2449, New York, NY, USA. Association for Computing
  Machinery.

\bibitem[{Thapliyal et~al.(2022)Thapliyal, Pont-Tuset, Chen, and
  Soricut}]{thapliyal_crossmodal-3600_2022}
Ashish~V. Thapliyal, Jordi Pont-Tuset, Xi~Chen, and Radu Soricut. 2022.
\newblock \href {https://aclanthology.org/2022.emnlp-main.45} {Crossmodal-3600:
  {A} {Massively} {Multilingual} {Multimodal} {Evaluation} {Dataset}}.
\newblock In \emph{Proceedings of the 2022 {Conference} on {Empirical}
  {Methods} in {Natural} {Language} {Processing}, {EMNLP} 2022, {Abu} {Dhabi},
  {United} {Arab} {Emirates}, {December} 7-11, 2022}, pages 715--729.
  Association for Computational Linguistics.

\bibitem[{Visheratin(2023)}]{visheratin_nllb_2023}
Alexander~A. Visheratin. 2023.
\newblock \href {https://doi.org/10.48550/ARXIV.2309.01859} {{NLLB-CLIP} -
  train performant multilingual image retrieval model on a budget}.
\newblock \emph{CoRR}, abs/2309.01859.

\bibitem[{Vrande{\v{c}}i{\'c}(2012)}]{vrandevcic2012wikidata}
Denny Vrande{\v{c}}i{\'c}. 2012.
\newblock Wikidata: A new platform for collaborative data collection.
\newblock In \emph{Proceedings of the 21st international conference on world
  wide web}, pages 1063--1064.

\bibitem[{Wehrmann et~al.(2019)Wehrmann, Lopes, Souza, and
  Barros}]{wehrmann_language-agnostic_2019}
Jonatas Wehrmann, Maurício~Armani Lopes, Douglas~M. Souza, and Rodrigo~C.
  Barros. 2019.
\newblock \href {https://doi.org/10.1109/ICCV.2019.00590} {Language-{Agnostic}
  {Visual}-{Semantic} {Embeddings}}.
\newblock In \emph{2019 {IEEE}/{CVF} {International} {Conference} on {Computer}
  {Vision}, {ICCV} 2019, {Seoul}, {Korea} ({South}), {October} 27 - {November}
  2, 2019}, pages 5803--5812. IEEE.

\bibitem[{Yang et~al.(2022)Yang, Pan, Lin, Men, Zhang, Zhou, and
  Zhou}]{yang_chinese_2022}
An~Yang, Junshu Pan, Junyang Lin, Rui Men, Yichang Zhang, Jingren Zhou, and
  Chang Zhou. 2022.
\newblock \href {https://doi.org/10.48550/arXiv.2211.01335} {Chinese {CLIP}:
  {Contrastive} {Vision}-{Language} {Pretraining} in {Chinese}}.
\newblock \emph{CoRR}, abs/2211.01335.
\newblock ArXiv: 2211.01335.

\bibitem[{Yoshikawa et~al.(2017)Yoshikawa, Shigeto, and
  Takeuchi}]{yoshikawa_stair_2017}
Yuya Yoshikawa, Yutaro Shigeto, and Akikazu Takeuchi. 2017.
\newblock \href {https://doi.org/10.18653/v1/P17-2066} {{STAIR} {Captions}:
  {Constructing} a {Large}-{Scale} {Japanese} {Image} {Caption} {Dataset}}.
\newblock In \emph{Proceedings of the 55th {Annual} {Meeting} of the
  {Association} for {Computational} {Linguistics} ({Volume} 2: {Short}
  {Papers})}, pages 417--421, Vancouver, Canada. Association for Computational
  Linguistics.

\bibitem[{Zeng et~al.(2022)Zeng, Zhou, Luo, and Zhang}]{zeng_cross-view_2022}
Yan Zeng, Wangchunshu Zhou, Ao~Luo, and Xinsong Zhang. 2022.
\newblock \href {https://doi.org/10.48550/arXiv.2206.00621} {Cross-{View}
  {Language} {Modeling}: {Towards} {Unified} {Cross}-{Lingual} {Cross}-{Modal}
  {Pre}-training}.
\newblock \emph{CoRR}, abs/2206.00621.
\newblock ArXiv: 2206.00621.

\bibitem[{Zhai et~al.(2023)Zhai, Mustafa, Kolesnikov, and
  Beyer}]{zhai_sigmoid_2023}
Xiaohua Zhai, Basil Mustafa, Alexander Kolesnikov, and Lucas Beyer. 2023.
\newblock \href {https://doi.org/10.48550/arXiv.2303.15343} {Sigmoid {Loss} for
  {Language} {Image} {Pre}-{Training}}.
\newblock \emph{CoRR}, abs/2303.15343.
\newblock ArXiv: 2303.15343.

\bibitem[{Zhai et~al.(2022)Zhai, Wang, Mustafa, Steiner, Keysers, Kolesnikov,
  and Beyer}]{zhai_lit_2022}
Xiaohua Zhai, Xiao Wang, Basil Mustafa, Andreas Steiner, Daniel Keysers,
  Alexander Kolesnikov, and Lucas Beyer. 2022.
\newblock \href {https://doi.org/10.1109/CVPR52688.2022.01759} {{LiT}:
  {Zero}-{Shot} {Transfer} with {Locked}-image text {Tuning}}.
\newblock In \emph{{IEEE}/{CVF} {Conference} on {Computer} {Vision} and
  {Pattern} {Recognition}, {CVPR} 2022, {New} {Orleans}, {LA}, {USA}, {June}
  18-24, 2022}, pages 18102--18112. IEEE.

\bibitem[{Zhang et~al.(2022)Zhang, Hu, and Jin}]{zhang_generalizing_2022}
Liang Zhang, Anwen Hu, and Qin Jin. 2022.
\newblock \href {https://doi.org/10.48550/arXiv.2206.11091} {Generalizing
  {Multimodal} {Pre}-training into {Multilingual} via {Language}
  {Acquisition}}.
\newblock \emph{CoRR}, abs/2206.11091.
\newblock ArXiv: 2206.11091.

\bibitem[{Zhou et~al.(2021)Zhou, Zhou, Wang, Cheng, Li, Yu, and
  Liu}]{zhou_uc2_2021}
Mingyang Zhou, Luowei Zhou, Shuohang Wang, Yu~Cheng, Linjie Li, Zhou Yu, and
  Jingjing Liu. 2021.
\newblock \href {https://doi.org/10.1109/CVPR46437.2021.00414} {{UC2}:
  {Universal} {Cross}-{Lingual} {Cross}-{Modal} {Vision}-and-{Language}
  {Pre}-{Training}}.
\newblock In \emph{{IEEE} {Conference} on {Computer} {Vision} and {Pattern}
  {Recognition}, {CVPR} 2021, virtual, {June} 19-25, 2021}, pages 4155--4165.
  Computer Vision Foundation / IEEE.

\end{thebibliography}
\bibliographystyle{acl_natbib}

\appendix

\section{License}
\label{sec:license}
\bin{} is a processed version of BabelNet v5.2 downloaded from \url{https://babelnet.org}, made available with the BabelNet Non-Commercial License (see \url{https://babelnet.org/full-license}).

%required statement:
% We bear all responsibility in case of violation of rights. 

\section{Data and Training Details}
\subsection{\bin}
\label{sec:appendix:detail:data}

\begin{table*}[t]
    \centering
    \footnotesize
    % \def\arraystretch{0.97}
    % \resizebox{0.98\linewidth}{!}{
%     \begin{tabular}{lr|lr|lr|lr|lr|lr|lr|lr|lr}
% af & 303 & am & 85 & ar & 636 & as & 98 & az & 365 & be & 415 & bg & 602 & bn & 282 & br & 297 \\
% bs & 156 & ca & 767 & cs & 615 & cy & 407 & da & 610 & de & 738 & el & 572 & eo & 603 & es & 845 \\
% et & 496 & eu & 625 & fa & 682 & fi & 973 & fr & 799 & fy & 155 & ga & 502 & gd & 217 & gl & 473 \\
% gu & 106 & ha & 47 & he & 648 & hi & 342 & hr & 347 & hu & 594 & hy & 454 & id & 463 & is & 409 \\
% it & 773 & ja & 733 & jv & 183 & ka & 438 & kk & 365 & km & 167 & kn & 175 & ko & 648 & ku & 101 \\
% ky & 247 & la & 276 & lo & 141 & lt & 535 & lv & 392 & mg & 64 & mk & 453 & ml & 281 & mn & 201 \\
% mr & 140 & ms & 419 & my & 232 & ne & 134 & nl & 749 & no & 599 & om & 18 & or & 71 & pa & 128 \\
% pl & 778 & ps & 112 & pt & 667 & ro & 687 & ru & 748 & sa & 66 & sd & 61 & si & 97 & sk & 509 \\
% sl & 393 & so & 58 & sq & 273 & sr & 468 & su & 98 & sv & 699 & sw & 220 & ta & 346 & te & 202 \\
% th & 896 & tl & 272 & tr & 559 & ug & 106 & uk & 640 & ur & 220 & uz & 254 & vi & 523 & xh & 35 \\
% yi & 175 & zh & 885 \\
    % \end{tabular}
    % [17, 32, 35, 16]
    \begin{tabular}{l p{12cm}}
\bf very low & \textit{om} (18), \textit{xh} (35), \textit{ha} (47), \textit{so} (58), \textit{sd} (61), \textit{nah} (62), \textit{hak} (63), \textit{mg} (64), \textit{sa} (66), \textit{or} (71), \textit{ce} (73), \textit{chr} (83), \textit{am} (85), \textit{diq} (89), \textit{si} (97), \textit{su} (98), \textit{as} (98) \\
\bf low &  \textit{ku} (101), \textit{gu} (106), \textit{ug} (106), \textit{ps} (112), \textit{pa} (128), \textit{ne} (134), \textit{cv} (137), \textit{mr} (140), \textit{lo} (141), \textit{fy} (155), \textit{bs} (156), \textit{km} (167), \textit{kn} (175), \textit{yi} (175), \textit{jv} (183), \textit{mn} (201), \textit{te} (202), \textit{gd} (217), \textit{sw} (220), \textit{ur} (220), \textit{my} (232), \textit{ky} (247), \textit{uz} (254), \textit{nv} (257), \textit{tl} (272), \textit{sq} (273), \textit{la} (276), \textit{wuu} (278), \textit{ml} (281), \textit{bn} (282), \textit{br} (297), \textit{af} (303) \\
\bf mid & \textit{hi} (342), \textit{ta} (346), \textit{hr} (347), \textit{az} (365), \textit{kk} (365), \textit{lv} (392), \textit{sl} (393), \textit{cy} (407), \textit{is} (409), \textit{be} (415), \textit{ms} (419), \textit{ka} (438), \textit{mk} (453), \textit{hy} (454), \textit{id} (463), \textit{sr} (468), \textit{gl} (473), \textit{et} (496), \textit{ga} (502), \textit{sk} (509), \textit{vi} (523), \textit{lt} (535), \textit{tr} (559), \textit{el} (572), \textit{hu} (594), \textit{no} (599), \textit{bg} (602), \textit{eo} (603), \textit{da} (610), \textit{cs} (615), \textit{eu} (625), \textit{ar} (636), \textit{uk} (640), \textit{ko} (648), \textit{he} (648) \\
\bf high &  \textit{pt} (667), \textit{fa} (682), \textit{ro} (687), \textit{sv} (699), \textit{ja} (733), \textit{de} (738), \textit{ru} (748), \textit{nl} (749), \textit{ca} (767), \textit{it} (773), \textit{pl} (778), \textit{fr} (799), \textit{es} (845), \textit{zh} (885), \textit{th} (896), \textit{fi} (973)
\end{tabular}
    % }
    % \caption{The 92 languages of Babel-ImageNet in alphabetical order with the corresponding number of classes in \bin{}.}
        \caption{The 100 languages of Babel-ImageNet in their respective groups. Number of classes in parentheses.}
    \label{tab:data:lang_classes}
\end{table*}

\begin{figure*}[t]
    \centering
    \includegraphics[width=\textwidth]{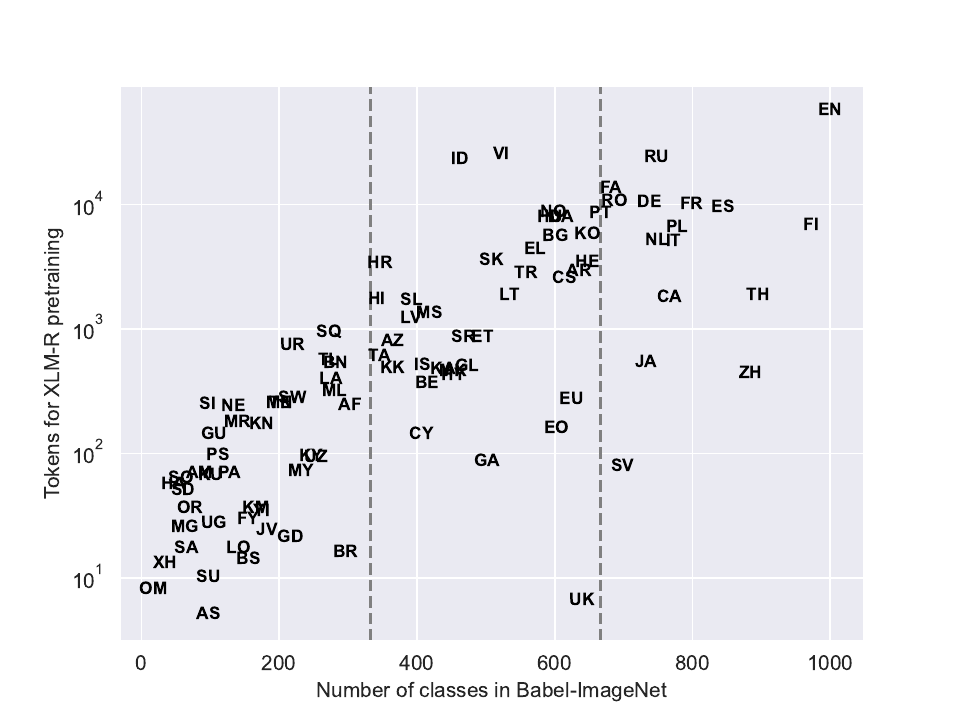}
    \caption{Number of classes in Babel-ImageNet plotted against the number of tokens (millions, log10) in the XLM-R pretraining corpus. When taking the XLM-R tokens as proxy for "resourceness" of a language, we see that this generally correlates with the number of classes. Vertical lines indicate the grouping of languages for evaluation.
    }
    %\vspace{-1mm}
\label{fig:data:class_resource}
%\vspace{-3mm}
\end{figure*}
Table \ref{tab:data:lang_classes} lists the 100 \bin{} languages with their corresponding number of classes.

Figure \ref{fig:data:class_resource} visualizes the relationship between the number of classes of a language in \bin{} and the number of tokens for the language in the XLM-R pretraining corpus (which we use as a proxy for the language ``resourceness''). We see that the two are generally correlated (Spearman rank correlation of 0.78), albeit with some expected outliers, e.g., Chinese is ''token-compact`` so the token count does not reflect its high-resourceness well.

\subsection{Prompts}
\label{sec:appendix:detail:prompts}
We use NLLB \citep{costa-jussa_no_2022} (\textit{nllb-200-distilled-1.3B}) to translate the 80 prompts used by \citet{radford_learning_2021} to our 100 languages. 
The exceptions are \textit{fy, la, br, wuu, nv, cv, diq, chr, ce, hak, nah}, which are not supported by NLLB; for those languages we report the better results between: (1) using only the language-specific labels and (2) inserting `labels' into English prompts. 
For the ISO 639-1 languages corresponding to macro languages, there is only one corresponding ISO 639-3 language in NLLB, except for \textit{no} where we choose Bokmål and for \textit{az} where we choose North Azerbaijani. We translate the prompts in their template form with the \textsc{\{\}} placeholders.
We use a range of different methods like HTML tags or other special characters to increase the likelihood of preserving the placeholders during translation and then select the first successful approach. If no method worked, we append \textsc{\{\}} to the end of the sentence.
We perform no language-specific adaptions like combining prompt variants with definite and indefinite articles for languages where this distinction does not exist (or articles do not exist at all) nor do we account for the grammatical gender of the classes when inserting them in the template.

\subsection{Training}
\label{sec:appendix:hyperparameters}

\textbf{Training Data:} For the language-specific adaptation training in \S\ref{sec:experiments:one_lang_train}, we leveraged the BLIP \citep{li_blip_2022} image-caption dataset \textsc{ccs\_synthetic\_filtered\_large.json}\footnote{\url{https://github.com/salesforce/BLIP\#pre-training-datasets-download}}.

\textbf{Hyperparameters:} We train with AdamW \citep{loshchilov_decoupled_2019}, 0.1 weight decay, a linear learning rate schedule with 20\% warmup, learning rate 1e-3 (chosen with sweep over 1e-3, 5e-4, 3e-4, 1e-4), batch size 512 (OpenCLIP)/ 192 (M-CLIP), for 100 epochs (OpenCLIP)/ 15 epochs (M-CLIP; longer training yielded no improvements).
Hyperparameters are chosen based on results on Sinhala.
We perform no early stopping and use the last epoch for evaluation.
The temperature for the contrastive loss is a trainable parameter as in \citet{radford_learning_2021} but we freeze it for text contrastive loss (training it resulted in worse results).
The maximum text sequence length is 70.
For adapters, we use the Pfeiffer architecture \citep{pfeiffer_mad-x_2020} (task adapters, not language adapters) with reduction factor 16 with the implementation from
AdapterHub \citep{pfeiffer_adapterhub_2020}.
We pre-encode images and English captions; i.e. the English embeddings for MSE and contrastive loss are not computed by the trained model but come from the model before training.
We do not use any type of image augmentation.

\textbf{Negative results:} We experimented with the following methods but did not pursue them further due to not-better or poor results. 
\begin{enumerate}
    \item Training with MSE loss using aligned English-X sentences from WikiMatrix \citep{schwenk_wikimatrix_2021}, similar to the ST and (in part) AltCLIP models, resulted in a performance decrease throughout (except for \textit{si} with OpenCLIP) as Table~\ref{tab:res:train_onelang_wikimatrix} shows. This suggests that it is important to use \textit{``visually-descriptive''} parallel data (i.e., parallel image captions), rather than \textit{any} parallel data.
    \item LoRA fine-tuning \citep{hu_lora_2022} ($\alpha=8, r=16$, lr 1e-3 after sweep) significantly ($>$10\% on \textit{si}) underperformed adapter-based fine-tuning.
    \item SimCSE loss (a self-supervised objective) \citep{gao_simcse_2021} based only on target-language captions yielded no improvements compare to the initial model, i.e., without any additional language-specialization training (experimented with OpenCLIP and batch size 256).
    \item Multitask training with both LiT and MSE distillation objectives produced no gains compared to training only with the MSE objective.
\end{enumerate}

\begin{table*}[t]
    \centering
    \footnotesize
     \def\arraystretch{0.97}
     \resizebox{\linewidth}{!}{
    \begin{tabular}{ll rrrrrrrrrrrrrrrr}
    \toprule
 \bf Model & \bf Loss & \bf xh & \bf si & \bf lo & \bf ur & \bf my & \bf hi & \bf ms & \bf et & \bf sk & \bf lt & \bf eu & \bf ar & \bf ko & \bf fa & \bf de & \bf zh\\
\midrule
M-CLIP XLMR-L B-32 & No training & 17.7 & 33.6 & 12.5 & 29.4 & 14.6 & 36.4 & 36.6 & 41.4 & 39.7 & 27.5 & 18.3 & 30.1 & 21.4 & 25.0 & 38.7 & 32.7 \\
 \cmidrule(lr){2-2}
 & MSE (WikiData) & --- & \cellcolor{red!15}25.0 & --- & --- & --- & \cellcolor{red!15}22.2 & --- & \cellcolor{red!15}21.3 & \cellcolor{red!15}23.3 & \cellcolor{red!15}21.8 & \cellcolor{orange!15}16.9 & \cellcolor{red!15}24.0 & \cellcolor{red!15}16.3 & \cellcolor{orange!15}22.5 & \cellcolor{red!15}31.4 & \cellcolor{red!15}24.3 \\
\midrule
OpenCLIP XLMR B-32 & No training & 24.4 & 3.1 & 0.7 & 25.8 & 5.8 & 25.8 & 37.4 & 29.8 & 45.1 & 35.2 & 17.1 & 24.6 & 33.8 & 32.7 & 47.8 & 40.9 \\
\cmidrule(lr){2-2}
& MSE (WikiData) & --- & \cellcolor{teal!15}17.1 & --- & --- & --- & \cellcolor{red!15}18.1 & --- & \cellcolor{red!15}18.5 & \cellcolor{red!15}25.0 & \cellcolor{red!15}23.0 & 17.7 & \cellcolor{orange!15}21.1 & \cellcolor{red!15}14.7 & \cellcolor{red!15}23.0 & \cellcolor{red!15}38.6 & \cellcolor{red!15}28.5 \\ \\
 \bottomrule
    \end{tabular}
    }
    \caption{Results of adapter-based language adaptation of M-CLIP and OpenCLIP with TextMSE loss using aligned sentences from WikiMatrix. Colors denote the size of change in performance w.r.t. original CLIP model: {\color{red}$\leq-5$}, {\color{orange}$\leq0$}, $\leq5$, {\color{cyan}$\leq10$}, {\color{teal}$\leq20$}, {\color{blue}$>20$} (best viewed in color).
    }
    \label{tab:res:train_onelang_wikimatrix}
\end{table*}

\section{Further Experiments and Analysis}
\subsection{Experimental Validation of Machine-translated Prompts}
\label{sec:experiments:prompts}
\begin{figure*}[]
    \centering
    \includegraphics[width=\linewidth]{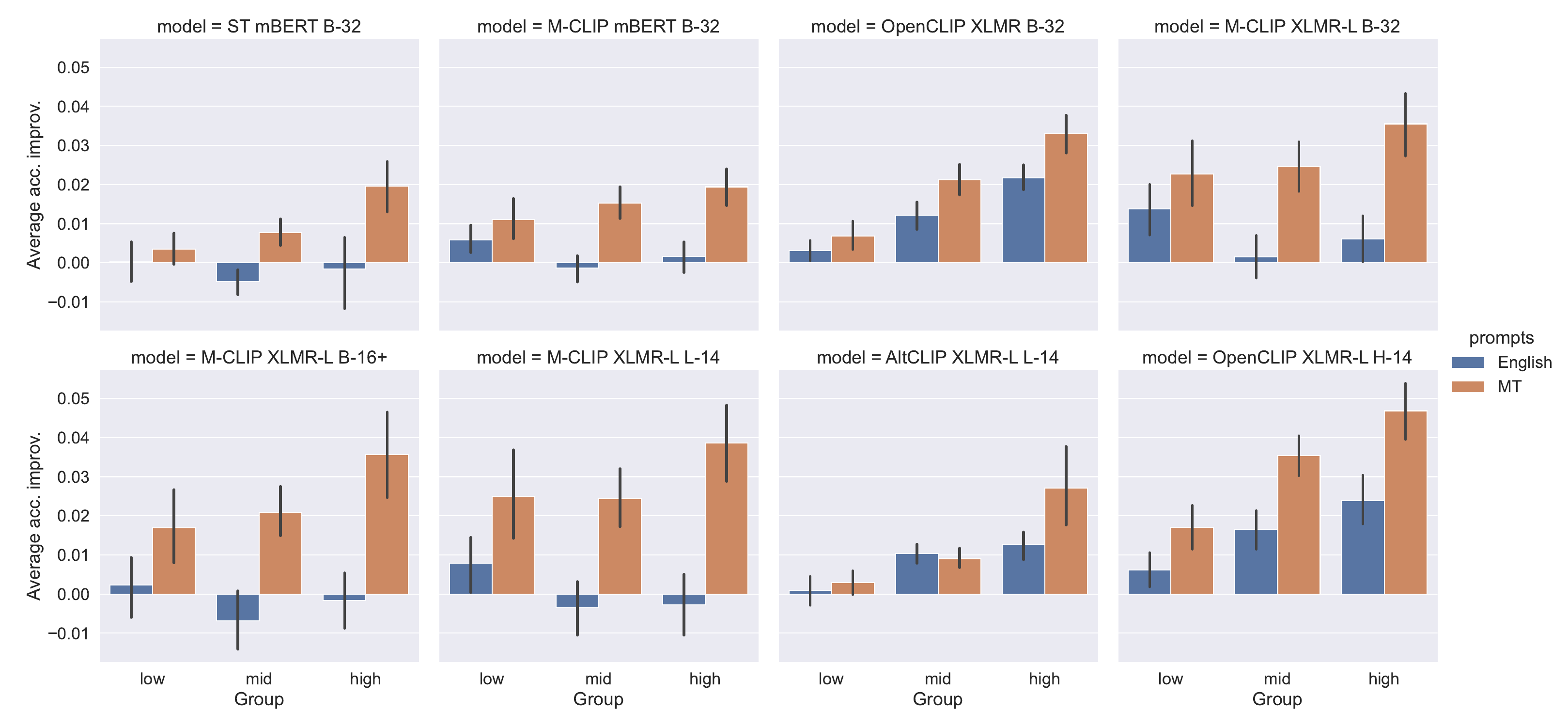}
    \caption{Average increase within the low/mid/high language groups (with 95\% CI) over only labels using English prompts (with non-English labels) and our machine-translated prompts.
    }
    %\vspace{-1mm}
\label{fig:plot:prompt_analysis}
%\vspace{-3mm}
\end{figure*}

We show in Figure \ref{fig:plot:prompt_analysis} that our translated prompts produce better results (on average, across all languages), compared to (i) using just the labels and (ii) inserting the translated labels into the  original English prompts created by \citet{radford_learning_2021}. With the translated prompts, we get gains of over 2 points for low-resource languages and up to 5 points for high-resource languages.

% We observe that 
%English prompts seem to work best for models trained  with substantial English data, i.e., the two OpenCLIP models trained on LAION5B where close to half the data is English. However, prompts in the same language as the labels still work better on average.
% Also, how much prompts improve over only using labels seems to depend on the performance on the language, i.e., if a model struggles on low-resource languages then using prompts will improve the performance less for them. 
% This is to be expected since the model's multilingual language capabilities affect both the zero-shot performance with only labels and also the ability to leverage the additional context from prompts.

\subsection{Comparison with Existing ImageNet Translations}
\label{sec:experiments:mt_imagenet}
\begin{figure*}[]
    \centering
    \includegraphics[width=\linewidth]{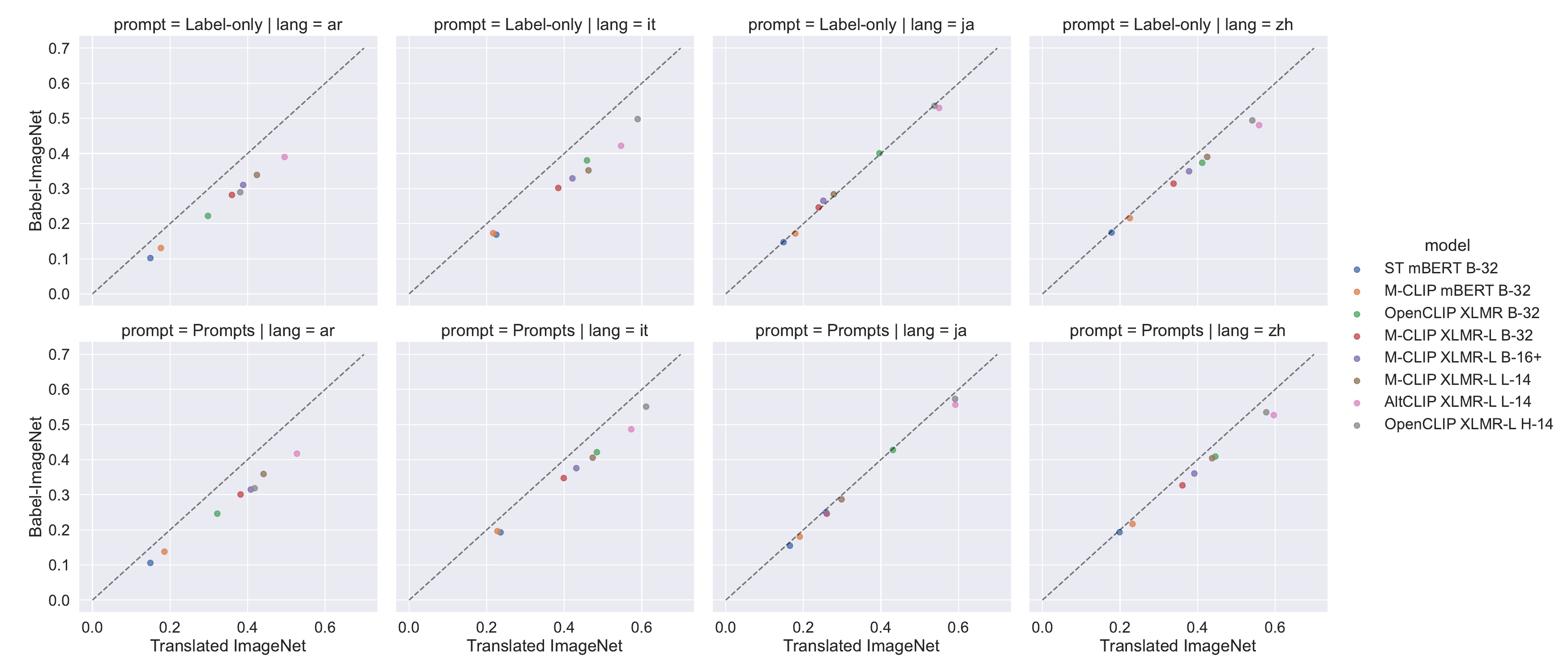}
    \caption{Results on \bin{} against results on existing ImageNet translations (``labels only'' in top row and with our MT prompts vs. human-created prompts in bottom row) for four languages: \textit{ar, zh, jp, it}. Relative model ranking is nearly identical between \bin{} and manually translated ImageNet benchmarks of respective languages.
    }
    %\vspace{-1mm}
\label{fig:plot:mt_imagenet}
%\vspace{-3mm}
\end{figure*}

% In appendix?
Prior work has created full translations of the 1k ImageNet classes into \textit{ar, zh, jp, it} along with human-written prompts for those languages.
We use those translations to validate our BabelNet-derived labels and MT prompts:
We evaluate models on the subset of ImageNet classes available for each language in \bin{} and compare a) only labels and b) human-created templates vs. our MT prompts.
Results are shown in Figure \ref{fig:plot:mt_imagenet}.
While results for \textit{ar} and \textit{it} are slighly higher in absolute numbers on the existing translation, the relative order of models on the \bin{} benchmarks of those languages is nearly identical to their relative ranking on the respective benchmarks with manually translated ImageNet labels.

We observe that the human-written prompts do not result in a relative improvement over our MT prompts (i.e. no down-shift parallel to the $x=y$ line). 
In fact, for \textit{it}, our MT prompts even close the gap slightly compared to the label-only setup.
% However, we should not conclude that MT prompts are always comparable to human-written ones as none of the four languages is considered low-resource where MT performance is poor.
% Hence, for low-resource languages, human-written prompts are likely to beat the MT prompts.

\subsection{Performance Differences between Distilled and Not-Distilled Languages}
\label{sec:experiments:distillation}
With teacher distillation, one would expect the performance in the languages seen in the distillation data to be better than in other languages, not used for distillation. With the wide language selection of our benchmark, we can analyze in-depth how performance on ``distilled'' languages differs from the performance on ``non-distilled'' languages.
% However, retrieval datasets with a limited range of (usually high-resource) languages are likely to be covered by the set of languages chosen for distillation\footnote{We note that 7/36 of XM3600 languages are not distilled by M-CLIP so we could also use it for this analysis but with only 36 languages in total, it is possible that another distilled CLIP model would cover all XM3600 languages.}.
% Even AltCLIP with only 9 languages covers most languages in xFlickrCo and XTD.

\begin{table*}[t]
    \centering
    \footnotesize
    % \def\arraystretch{0.97}
    % \resizebox{0.98\linewidth}{!}{
    \begin{tabular}{l rrrrrr}
    \toprule
\bf Model & low &  $\Delta$low &  mid &  $\Delta$mid &  high &  $\Delta$high \\
\midrule
 M-CLIP XLMR-L L-14 &        37.6 &     +14.4 &        45.9 &     +16.8 &        41.6 &     +11.3 \\
OpenCLIP XLMR-L H-14 &        26.1 &     +10.1 &        47.5 &     +13.2 &        55.3 &     +15.8 \\
  \cmidrule(lr){2-3} \cmidrule(lr){4-5} \cmidrule(lr){6-7}
AltCLIP XLMR-L L-14 &         --- &      --- &        47.5 &     +28.0 &        51.7 &     +28.9 \\
OpenCLIP XLMR-L H-14 &         --- &      --- &        37.8 &     -3.5 &        57.0 &      +7.4 \\
    \bottomrule
    \end{tabular}
    % }
    \caption{Average results for the ``distilled'' languages in the low/mid/high-resource language groups and the $\Delta$ difference to the other ``non-distilled'' languages of the groups. OpenCLIP H-14 serves as control for language-specific differences in performance not caused by distillation. For M-CLIP, 14/41, 18/35, and 13/16 languages per group are distilled; For AltCLIP, 0/41,  2/35, and 6/16 are distilled.}
    \label{tbl:analysis:distill}
\end{table*}

We compare results for distilled languages on the low/mid/high-resource language groups for M-CLIP and AltCLIP in Table \ref{tbl:analysis:distill}; we use the OpenCLIP H-14 model as reference for an expected `baseline' $\Delta$-difference in performance between the distilled/not-distilled language groups that is due to other factors inherent to the specific languages and not the distillation.
For AltCLIP, we see that the the performance on the 8 distilled languages is significantly better than on the non-distilled languages. Moreover, the performance on its distilled languages is even comparable to that of the larger H-14 model. For M-CLIP, the performance on the distilled languages is only slightly better than on the non-distilled low- and mid-resource languages when compared to the OpenCLIP model and the gap is even smaller for high-resource languages. Interestingly, the performance on non-distilled low-resource languages is still noticeably better for M-CLIP than for the OpenCLIP H-14 model. We speculate that the shorter training of M-CLIP compared to OpenCLIP might retain more of the language-specific competences for low-resource languages, obtained in XLM-R pretraining.

\subsection{LAION5B: Language Distribution and Performance}
\label{sec:experiments:lang_distribution}

\begin{figure}[ht!]
     \centering
     \begin{subfigure}[b]{0.45\textwidth}
         \centering
         \includegraphics[width=\textwidth]{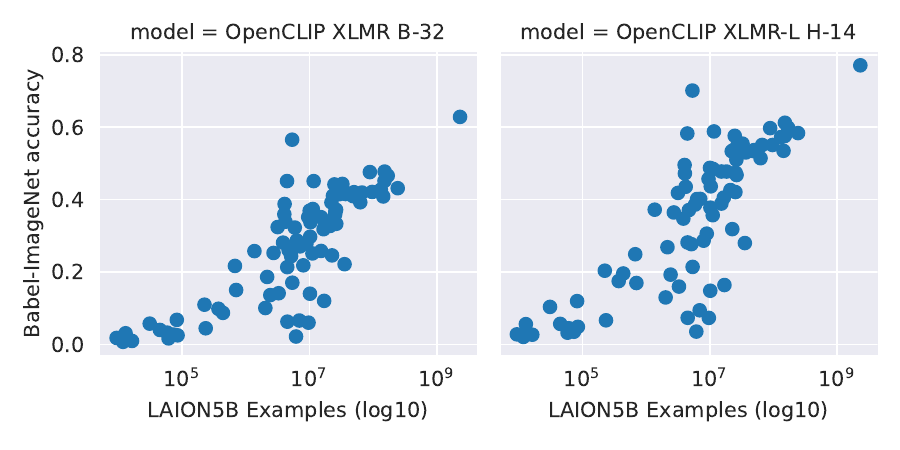}
         \caption{Accuracy of the LAION5B-trained OpenCLIP models for the 92 \bin{} languages plotted against the number of examples for each language in LAION5B (log10).}
         \label{fig:laion:accuracy}
     \end{subfigure}
     \begin{subfigure}[b]{0.45\textwidth}
         \centering
         \includegraphics[width=\textwidth]{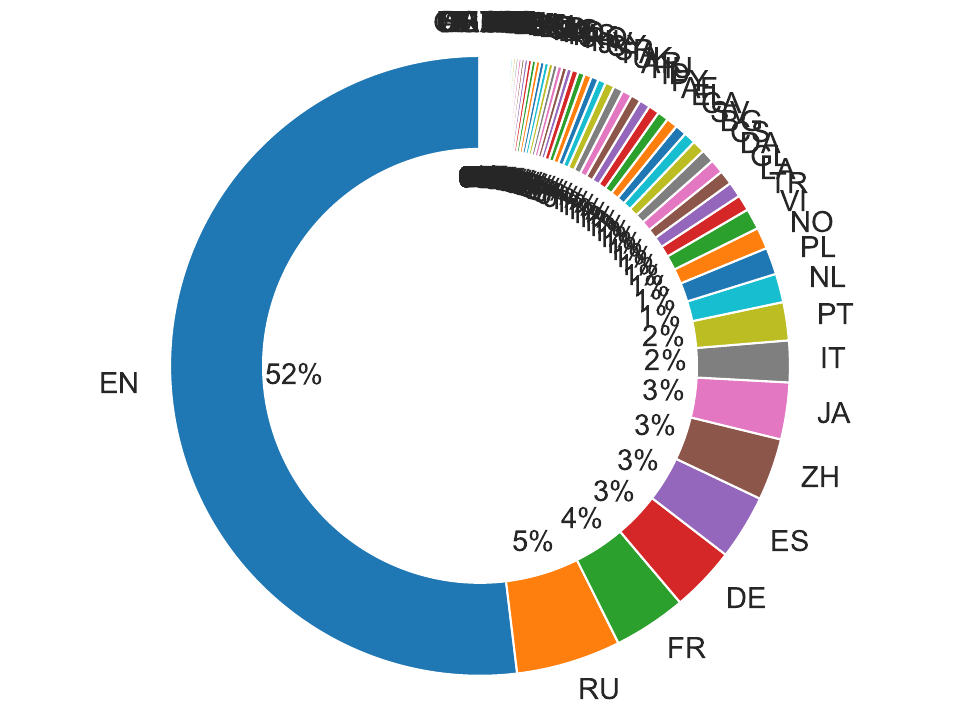}
         \caption{Distribution of languages in LAION5B (exluding the 1.3B no-language examples) shows that most examples are either English or one of a few other languages.}
         \label{fig:laion:distribution}
     \end{subfigure}
        \caption{The relationship between the LAION5B language distribution with the performance of OpenCLIP models trained on that data.}
        \label{fig:laion}
\end{figure}

The distillation-based models evaluated in our benchmark use the same number of training examples for every language.
The OpenCLIP models, on the other hand, are trained on LAION5B which follows a more `natural' distribution of image-caption pairs across languages, as found on the web: Figure \ref{fig:laion:distribution} shows that over half the data is English, 7 high-resource languages account for another 25\% of the data, whereas all remaining languages ``share'' the remaining 25\%.

We can see in Figure \ref{fig:laion:accuracy} that the OpenCLIP's ZS-IC performance for \bin{} languages highly depends on the number of instances of those languages in the LAION5B dataset. The Spearman rank correlation between the number of language-specific LAION5B examples and the respective \bin{} accuracy for the language is 0.76. This suggests that pure image-text contrastive pre-training results in poor generalization and limited cross-lingual gains to languages unseen in pretraining. Additional training objectives that aim to better align the multilingual space for example using paired text like in MURAL \citep{jain_mural_2021} might be necessary to improve results of OpenCLIP-like models (trained from scratch) for low-resource languages.

\newpage
\section{Full Results}
We report full results for all languages on (i) \bin{} and (ii) each of the three image-text retrieval datasets: xFlickrCo, XM3600, and XTD.
\begin{table*}
    \centering
    \footnotesize
    
     \def\arraystretch{0.97}
     \resizebox{1\linewidth}{!}{
\begin{tabular}{lrrrrrrrrrrrrr}

\toprule
% lang   &  OAI B32 &  ST B32 &  MC mB B32 &  OC B32 & MC B32 &  MC B16+ &  MC L14 &  AC L14 &  OC H14 \\
lang &  AC L-14 &  MC mB B-32 &  MC B-16+ &  MC B-32 &  MC L-14 &  NS-base &  NS-large &  OAI B-32 &  OC B-32 &  OC H-14 &  mSigLIP &  ST B-32 \\
\midrule
en    &                73.36 &              29.97 &                47.02 &               44.06 &               52.34 &             39.75 &              51.96 &        63.35 &               62.32 &                 76.95 &    75.12 &          39.45 \\
af    &                 22.6 &              25.26 &                50.33 &               47.48 &               55.32 &             41.11 &              53.89 &        10.65 &               36.48 &                 47.53 &    47.17 &          14.94 \\
am    &                 7.13 &               0.99 &                27.22 &               29.25 &               29.11 &             50.12 &              55.79 &         1.51 &                1.04 &                  2.56 &    14.56 &           0.85 \\
ar    &                42.44 &              13.99 &                32.06 &               30.68 &                36.3 &             29.52 &               40.1 &         0.56 &               24.51 &                  31.7 &    42.27 &          10.69 \\
as    &                 6.53 &              17.22 &                21.69 &               22.84 &               24.92 &             39.94 &              49.22 &         1.18 &                5.43 &                  9.53 &    23.14 &           2.92 \\
az    &                 16.4 &              23.71 &                26.57 &               25.69 &               28.81 &             34.77 &              44.24 &         7.27 &               25.83 &                 36.98 &    47.82 &          12.68 \\
be    &                27.52 &              14.08 &                35.35 &               32.66 &               36.85 &             36.72 &               48.4 &         0.64 &               27.18 &                 40.12 &    32.05 &          10.02 \\
bg    &                34.48 &              21.61 &                42.63 &               40.57 &               47.13 &             32.23 &              42.43 &          0.8 &               41.26 &                 54.02 &    58.99 &          19.56 \\
bn    &                 7.25 &              25.99 &                31.17 &               29.77 &               35.55 &             47.48 &              57.68 &         0.17 &                8.61 &                 19.57 &    49.42 &           3.06 \\
br    &                12.02 &               3.99 &                10.71 &                9.64 &               11.61 &              8.82 &              12.28 &         7.14 &               13.02 &                 15.23 &    14.96 &           3.47 \\
bs    &                29.45 &              39.15 &                60.18 &               59.74 &               65.27 &             49.55 &              58.69 &        13.04 &               55.99 &                 69.88 &    79.49 &          35.35 \\
ca    &                32.02 &               19.9 &                37.48 &               34.94 &               40.33 &             28.34 &              38.67 &        11.54 &               33.37 &                 46.78 &    42.96 &           17.5 \\
ce    &                16.38 &                6.9 &                15.73 &               11.37 &               13.53 &             19.04 &              20.22 &         1.18 &               21.21 &                 22.19 &    25.53 &           4.36 \\
chr   &                  1.2 &                1.2 &                  1.2 &                1.16 &                1.47 &              1.18 &               2.19 &         1.47 &                1.33 &                  0.48 &     1.54 &           1.64 \\
cs    &                20.03 &              22.04 &                38.44 &               36.59 &               41.89 &             31.26 &              42.06 &         6.36 &                41.8 &                 54.11 &    66.02 &           19.5 \\
cv    &                29.93 &              11.75 &                23.64 &               19.78 &               22.99 &             17.55 &              24.67 &         0.58 &                31.8 &                 34.67 &    32.96 &          10.32 \\
cy    &                 12.8 &              19.81 &                11.39 &               11.31 &               12.99 &              25.8 &              33.45 &         5.04 &               12.08 &                 16.52 &    12.72 &           5.62 \\
da    &                24.96 &              22.79 &                46.92 &               44.45 &               51.84 &             34.66 &              46.87 &        12.39 &               42.92 &                  55.5 &    62.46 &          22.65 \\
de    &                27.96 &              20.07 &                42.01 &               39.59 &               46.39 &             31.74 &              44.43 &        15.83 &               47.58 &                 61.05 &     67.1 &           18.7 \\
diq   &                14.81 &               6.49 &                20.67 &               19.19 &               18.47 &             17.62 &              22.11 &          8.2 &               18.72 &                 21.42 &    25.96 &           9.08 \\
el    &                 5.03 &              16.58 &                41.73 &               39.03 &               45.75 &              33.3 &              45.72 &          0.9 &               37.21 &                 50.77 &    51.16 &          15.29 \\
eo    &                24.84 &              13.04 &                25.22 &               24.64 &               28.58 &             35.12 &              48.63 &         5.77 &               21.82 &                  28.5 &    21.65 &           10.9 \\
es    &                51.78 &              20.13 &                38.64 &               36.31 &               42.07 &             32.14 &              42.42 &        16.98 &               44.78 &                  57.4 &    60.72 &          20.67 \\
et    &                15.54 &              19.79 &                45.19 &               42.22 &               48.73 &             29.21 &              39.91 &         4.64 &               29.68 &                 37.84 &     53.5 &          13.51 \\
eu    &                17.72 &               9.63 &                18.92 &               18.72 &               21.25 &             34.16 &              45.22 &         7.43 &               16.87 &                 21.51 &    21.89 &            9.0 \\
fa    &                20.13 &              20.69 &                28.24 &               25.55 &               29.26 &              27.6 &              37.34 &         0.46 &               32.66 &                 42.48 &     52.3 &          13.89 \\
fi    &                10.56 &              14.45 &                27.49 &               25.79 &               30.27 &             22.82 &              32.19 &         3.96 &                25.0 &                 35.69 &    46.88 &          11.25 \\
fr    &                54.06 &              20.17 &                38.84 &               35.84 &               42.39 &             32.21 &              43.05 &        21.08 &               46.21 &                 59.72 &    63.54 &          20.93 \\
fy    &                23.73 &               6.83 &                 27.9 &               28.72 &               30.09 &              22.7 &              29.64 &        11.99 &               34.13 &                 40.65 &    36.21 &           6.15 \\
ga    &                 8.36 &               3.12 &                 9.42 &                8.29 &                9.76 &             20.12 &              27.98 &         2.72 &                6.71 &                  9.45 &     5.55 &           2.24 \\
gd    &                 6.01 &               3.46 &                 8.25 &                7.52 &               10.53 &             25.43 &              30.48 &         3.05 &                5.98 &                   7.3 &     4.94 &           2.63 \\
gl    &                43.92 &              23.08 &                41.53 &               40.26 &               45.05 &             35.13 &              47.57 &        15.47 &               43.82 &                 54.66 &    47.37 &          23.91 \\
gu    &                11.34 &              28.77 &                28.11 &               26.94 &               32.32 &              49.3 &              63.42 &          0.4 &                6.94 &                 12.02 &    20.72 &           7.64 \\
ha    &                 3.23 &              19.36 &                 4.94 &                4.17 &                5.19 &             41.96 &              49.74 &         1.62 &                2.47 &                  3.57 &     3.06 &            2.0 \\
hak   &                33.27 &               18.7 &                32.29 &               26.03 &                28.1 &             21.94 &               27.9 &         3.97 &                32.6 &                 35.08 &    31.14 &          11.08 \\
he    &                 7.03 &              16.68 &                22.04 &               21.02 &                23.7 &             26.07 &              35.16 &          0.3 &               26.64 &                 35.87 &    48.99 &          12.86 \\
hi    &                12.89 &              20.99 &                37.37 &               37.04 &               41.61 &             41.47 &              54.75 &         0.09 &               25.85 &                 38.65 &    40.51 &          17.58 \\
hr    &                21.75 &              30.22 &                51.08 &               50.67 &                57.0 &             40.19 &              51.75 &         8.63 &               44.82 &                  58.2 &    66.58 &          25.64 \\
hu    &                16.55 &              19.91 &                45.47 &               42.69 &                50.3 &             31.38 &              42.39 &         5.57 &               40.93 &                 53.25 &     63.9 &           17.1 \\
hy    &                  5.3 &              16.89 &                18.65 &               17.97 &               18.26 &              26.7 &               36.7 &         0.17 &                9.91 &                  17.5 &    35.09 &           7.43 \\
id    &                26.52 &               26.6 &                50.39 &               47.73 &               55.14 &             42.63 &              53.63 &        16.09 &               43.78 &                 57.61 &    69.25 &          23.49 \\
is    &                10.82 &              18.91 &                41.84 &               39.96 &               45.56 &             29.15 &              38.46 &         2.93 &               13.51 &                 19.25 &    30.18 &           6.04 \\
it    &                49.43 &              19.97 &                37.85 &               35.81 &               41.27 &             29.95 &              41.32 &        15.55 &               41.88 &                 55.03 &    59.89 &          19.49 \\
ja    &                56.33 &              18.55 &                25.57 &               24.93 &                29.2 &             28.91 &              37.71 &         4.18 &               42.35 &                 57.14 &    62.79 &          15.74 \\
jv    &                 21.8 &              19.51 &                33.85 &               32.89 &               37.27 &             42.74 &               52.2 &        14.89 &               31.81 &                 40.44 &     50.4 &          15.08 \\
ka    &                  9.0 &              15.48 &                22.83 &               22.05 &               24.26 &             26.69 &              34.27 &          0.2 &               11.09 &                 20.44 &    41.13 &           8.66 \\
kk    &                31.61 &              19.96 &                26.59 &               25.59 &               28.44 &             34.44 &              44.46 &         0.53 &               28.12 &                 34.52 &     47.7 &           10.8 \\
km    &                 6.72 &               0.91 &                16.57 &               17.68 &               19.77 &             25.78 &              30.16 &         0.79 &                3.31 &                  3.16 &    31.54 &           0.38 \\
kn    &                10.97 &              21.17 &                27.09 &               27.36 &               29.57 &              43.3 &              53.95 &         1.04 &                4.16 &                  5.68 &     16.3 &           2.16 \\
ko    &                 53.8 &              16.02 &                24.02 &               21.61 &               25.12 &             24.48 &              33.15 &         0.42 &               33.52 &                 43.51 &    56.58 &          12.52 \\
ku    &                11.23 &               8.95 &                12.75 &                14.2 &               14.79 &             14.73 &              21.27 &         5.09 &                14.1 &                 15.88 &     15.5 &           8.73 \\
ky    &                32.11 &              18.48 &                28.79 &                28.7 &               31.69 &             38.53 &              47.82 &         0.65 &               32.56 &                 38.55 &     41.6 &          14.01 \\
la    &                23.06 &                4.7 &                16.43 &               11.23 &               14.38 &              12.8 &              21.94 &        10.77 &               21.14 &                 26.65 &    26.35 &            4.2 \\
lo    &                 9.86 &               0.74 &                12.45 &               12.75 &               13.05 &             32.13 &              37.59 &         0.77 &                0.85 &                  2.21 &    10.26 &           0.71 \\
lt    &                 18.7 &              21.69 &                 29.9 &               28.06 &               31.96 &             29.09 &              39.78 &         5.94 &               34.68 &                 45.79 &    55.12 &          18.58 \\
lv    &                21.42 &              26.01 &                 33.9 &               34.14 &               37.14 &             31.99 &              40.98 &         7.74 &               38.52 &                  47.3 &     56.7 &          23.65 \\
mg    &                13.25 &               8.62 &                13.06 &               14.19 &               14.12 &             38.47 &              47.72 &         4.06 &               13.94 &                 14.62 &    12.34 &           9.06 \\
mk    &                31.41 &              23.55 &                47.78 &               46.53 &                51.4 &             33.18 &              43.89 &         1.33 &               36.08 &                 49.47 &    47.31 &          20.45 \\
ml    &                 7.64 &              20.05 &                38.69 &                38.8 &               44.01 &             37.69 &              49.56 &         0.16 &                1.69 &                  4.45 &    15.31 &           1.23 \\
mn    &                20.44 &              28.29 &                20.86 &               19.37 &               21.78 &             25.09 &               28.6 &         1.02 &               18.58 &                 26.49 &    44.96 &          14.06 \\
mr    &                17.46 &               25.9 &                45.89 &               45.41 &               47.83 &             59.39 &              68.63 &         1.13 &               32.29 &                 41.66 &    42.24 &          29.66 \\
ms    &                21.25 &              22.12 &                39.23 &               37.86 &               43.59 &             35.71 &              46.72 &        14.94 &               37.08 &                 48.35 &     61.1 &          19.26 \\
my    &                 9.47 &               2.49 &                14.72 &               14.49 &                16.2 &             38.34 &              46.23 &         0.49 &                5.82 &                 10.27 &    17.16 &          10.35 \\
nah   &                 8.13 &               5.52 &                 6.52 &                7.42 &                9.03 &              7.13 &               7.97 &         4.35 &               10.68 &                 11.35 &      9.9 &           2.23 \\
ne    &                16.51 &              17.07 &                37.57 &               40.16 &               39.21 &             49.78 &              60.39 &         0.54 &                25.0 &                 36.39 &    40.16 &           16.9 \\
nl    &                25.03 &              18.57 &                39.09 &               37.33 &               44.56 &             29.67 &              41.41 &        13.05 &               41.71 &                 54.93 &    57.96 &          18.88 \\
no    &                23.47 &              21.13 &                 43.4 &               40.66 &               47.56 &             33.24 &              43.78 &          9.4 &               41.98 &                 53.57 &    61.21 &          19.55 \\
nv    &                 0.65 &               0.39 &                  0.6 &                0.38 &                0.16 &              0.66 &               0.29 &         0.35 &                0.44 &                  0.41 &     0.25 &           0.54 \\
om    &                  8.0 &               3.44 &                11.11 &                9.67 &                9.67 &             29.22 &              30.44 &         4.44 &                5.44 &                  13.0 &    10.22 &           8.78 \\
or    &                12.62 &               1.75 &                30.62 &               31.18 &               36.11 &             57.24 &              70.56 &         1.75 &                3.27 &                  1.61 &     1.61 &           1.77 \\

\bottomrule
\end{tabular}
    }
    \caption{Babel-ImageNet results for all languages (sorted alphabetically expect for English). To save space, we shorten sources (OAI: OpenAI; OC: OpenCLIP; MC: M-CLIP; ST: SentenceTransformer, AC: AltCLIP, NS: NLLB-SigLIP) and remove the text model if possible.
    \label{tab:res:full}
    }
\end{table*}
\begin{table*}
    \centering
    \footnotesize
    
     \def\arraystretch{0.97}
     \resizebox{1\linewidth}{!}{
\begin{tabular}{lrrrrrrrrrrrrr}

\toprule
% lang   &  OAI B32 &  ST B32 &  MC mB B32 &  OC B32 & MC B32 &  MC B16+ &  MC L14 &  AC L14 &  OC H14 \\
lang &  AC L-14 &  MC mB B-32 &  MC B-16+ &  MC B-32 &  MC L-14 &  NS-base &  NS-large &  OAI B-32 &  OC B-32 &  OC H-14 &  mSigLIP &  ST B-32 \\
\midrule
pa    &                11.42 &               7.97 &                33.89 &               32.09 &               31.69 &             60.27 &              70.47 &         1.53 &                1.77 &                  2.56 &     3.05 &           4.09 \\
pl    &                19.72 &              18.34 &                35.83 &               33.33 &               38.95 &             28.89 &              39.18 &         7.63 &               39.15 &                 51.46 &    61.38 &          16.32 \\
ps    &                18.05 &              10.86 &                19.48 &               18.98 &               23.21 &             33.38 &               41.8 &         1.57 &               21.66 &                 25.04 &    24.52 &          11.88 \\
pt    &                37.41 &              21.57 &                43.24 &               40.06 &               47.36 &             34.42 &              46.26 &        14.43 &               47.21 &                 59.62 &     61.7 &          23.03 \\
ro    &                23.92 &              18.78 &                40.45 &               37.78 &               44.26 &             29.74 &              41.21 &        10.19 &               35.17 &                  47.6 &    51.22 &          16.22 \\
ru    &                48.87 &              18.34 &                36.82 &               35.25 &                41.7 &             30.47 &              41.96 &         0.54 &               42.95 &                 58.21 &    60.16 &          16.39 \\
sa    &                11.18 &               9.45 &                17.36 &               17.21 &               18.97 &             32.36 &              37.09 &         0.91 &               11.55 &                 11.73 &    23.15 &           9.55 \\
sd    &                15.93 &              20.98 &                28.13 &                27.9 &               29.87 &             43.34 &              51.84 &          2.0 &               14.56 &                 16.49 &    15.61 &           9.41 \\
si    &                 19.4 &               2.19 &                33.32 &               33.75 &               36.39 &             55.38 &              65.48 &         2.64 &                3.22 &                  5.32 &    29.13 &           2.12 \\
sk    &                22.97 &              25.31 &                43.22 &               40.56 &               46.11 &             35.56 &              46.68 &         8.65 &               44.99 &                 58.68 &    66.17 &          22.72 \\
sl    &                17.39 &              24.61 &                45.05 &               42.98 &               48.71 &             34.15 &              47.76 &         6.01 &               36.93 &                 48.65 &    59.23 &          22.46 \\
so    &                 3.76 &              17.55 &                10.38 &               12.55 &               12.14 &             29.97 &              38.86 &         2.48 &                6.66 &                  7.41 &     7.07 &           5.17 \\
sq    &                21.71 &              26.21 &                49.82 &               48.12 &               54.07 &             38.39 &              48.67 &         8.24 &               33.99 &                 43.35 &    43.77 &          24.86 \\
sr    &                30.64 &              19.18 &                45.99 &               44.58 &               49.92 &             32.35 &              44.06 &         1.18 &               33.93 &                  47.7 &    45.43 &          17.49 \\
su    &                18.73 &              16.04 &                28.16 &               29.22 &               29.33 &             37.39 &              44.04 &        12.06 &               27.71 &                 30.61 &     35.9 &          11.67 \\
sv    &                21.06 &               21.1 &                45.72 &                42.0 &               49.44 &             31.93 &              42.26 &         9.17 &               42.88 &                  55.2 &    61.67 &          19.68 \\
sw    &                10.37 &               17.3 &                37.05 &               36.85 &               39.38 &             31.87 &              38.76 &         4.72 &               10.25 &                 12.91 &    20.17 &           6.63 \\
ta    &                 5.53 &              18.67 &                18.78 &               18.09 &               20.98 &             30.98 &              41.14 &         0.35 &                4.54 &                  6.73 &    30.21 &           2.12 \\
te    &                10.75 &              24.95 &                32.56 &               31.75 &               34.44 &             45.62 &              58.13 &         0.51 &                2.83 &                  3.71 &    19.17 &           3.02 \\
th    &                11.85 &              15.55 &                29.69 &               27.77 &               32.77 &             25.63 &              33.35 &         1.33 &               28.71 &                 40.16 &    40.84 &          10.62 \\
tl    &                18.25 &              16.17 &                33.43 &               32.51 &               38.11 &             26.43 &              33.62 &         7.86 &               17.54 &                 21.73 &    33.05 &           8.18 \\
tr    &                17.67 &              23.16 &                44.98 &               42.78 &                48.4 &             35.84 &              46.72 &         7.61 &               41.25 &                 52.88 &    63.65 &          19.47 \\
ug    &                 9.06 &               2.17 &                13.19 &               11.68 &               12.75 &             33.42 &               40.3 &         1.19 &                3.06 &                  4.64 &     6.19 &            2.3 \\
uk    &                34.82 &              17.61 &                36.67 &               35.77 &               41.22 &             28.99 &               39.5 &         0.63 &               39.19 &                 53.18 &    57.54 &          15.13 \\
ur    &                18.48 &              27.24 &                27.15 &               29.85 &               31.17 &             47.62 &              58.92 &         0.59 &               26.01 &                 37.12 &    29.48 &          17.75 \\
uz    &                17.24 &               21.4 &                22.57 &               22.54 &               24.13 &             36.09 &              48.54 &         6.13 &               21.83 &                 28.45 &    36.42 &            9.8 \\
vi    &                11.69 &              19.09 &                39.59 &               37.94 &               44.58 &             29.77 &              39.13 &         6.94 &               40.92 &                 53.39 &     59.9 &          17.86 \\
wuu   &                72.53 &               24.4 &                49.64 &               28.58 &               48.78 &             31.35 &              41.76 &         3.43 &               59.02 &                 70.81 &    72.22 &          19.69 \\
xh    &                21.77 &              16.34 &                 19.2 &               17.71 &               20.11 &              57.6 &               69.2 &        18.86 &               24.11 &                 27.14 &     24.8 &          14.23 \\
yi    &                 5.29 &               1.04 &                18.37 &                18.9 &               19.38 &             39.22 &              52.81 &         0.71 &                2.49 &                  4.73 &     3.82 &           1.33 \\
zh    &                53.35 &              21.99 &                36.41 &                33.5 &                40.9 &              25.5 &              33.26 &         1.85 &               40.73 &                 53.28 &    55.42 &          19.75 \\
\bottomrule
\end{tabular}
    }
    \label{tab:res:full2}
    \caption{Babel-ImageNet results for all languages (sorted alphabetically expect for English). To save space, we shorten sources (OAI: OpenAI; OC: OpenCLIP; MC: M-CLIP; ST: SentenceTransformer, AC: AltCLIP) and remove the text model if possible.
    }
\end{table*}

\begin{table*}[]
    \centering
    \footnotesize
     \def\arraystretch{0.97}
     \resizebox{1\linewidth}{!}{
\begin{tabular}{lrrrrrrrrr}
\toprule
            model &     en &     de &     es &     id &     jp &     ru &     tr &     zh &  average \\
\midrule
 AltCLIP XLMR-L L-14 & 64.50 & 33.05 & 66.65 & 20.75 & 57.05 & 66.00 & 10.65 & 62.50 &    45.24 \\
   M-CLIP mBERT B-32 & 42.45 & 32.20 & 39.80 & 33.15 & 31.10 & 39.10 & 36.05 & 36.60 &    35.43 \\
 M-CLIP XLMR-L B-16+ & 63.80 & 59.10 & 67.35 & 62.75 & 50.45 & 72.35 & 66.25 & 63.20 &    63.06 \\
  M-CLIP XLMR-L B-32 & 49.15 & 42.70 & 48.55 & 43.85 & 33.50 & 51.60 & 47.00 & 44.65 &    44.55 \\
  M-CLIP XLMR-L L-14 & 58.00 & 50.85 & 58.45 & 52.95 & 42.40 & 59.00 & 55.85 & 52.10 &    53.09 \\
    NLLB-SigLIP-base & 66.80 & 58.30 & 67.45 & 60.65 & 53.10 & 70.65 & 64.90 & 59.25 &    62.04 \\
   NLLB-SigLIP-large & 70.55 & 65.00 & 72.30 & 65.10 & 60.20 & 74.55 & 68.90 & 63.20 &    67.04 \\
  OpenCLIP XLMR B-32 & 61.80 & 53.15 & 61.45 & 48.90 & 47.90 & 65.50 & 53.00 & 54.95 &    54.98 \\
OpenCLIP XLMR-L H-14 & 73.85 & 66.85 & 77.65 & 64.80 & 63.70 & 78.60 & 68.85 & 69.90 &    70.05 \\
             mSigLIP & 68.45 & 59.15 & 70.45 & 57.50 & 29.45 & 71.75 & 55.35 & 47.80 &    55.92 \\
       ST mBERT B-32 & 39.85 & 24.35 & 29.95 & 26.65 & 20.40 & 28.15 & 22.50 & 26.20 &    25.46 \\
\bottomrule
\end{tabular}
}
    \caption{xFlickrCo T2I R@1. Average is without English.}
    \label{tab:res:xflickrco}
\end{table*}

\begin{table*}[]
    \centering    
    \footnotesize
     \def\arraystretch{0.97}
     \resizebox{1\linewidth}{!}{
\begin{tabular}{lrrrrrrrrrrr}
\toprule
% lang   &  OAI B32 &  ST B32 &  MC mB B32 &  OC B32 & MC B32 &  MC B16+ &  MC L14 &  AC L14 &  OC H14 \\
lang &  AC L-14 &  MC mB B-32 &  MC B-16+ &  MC B-32 &  MC L-14 &  NS-base &  NS-large &   OC B-32 &  OC H-14 &  mSigLIP &  ST B-32 \\
\midrule
en      &                43.49 &              24.63 &                47.47 &               31.69 &               36.56 &             48.25 &              50.14 &               48.38 &                 54.43 &    52.17 &          25.58 \\
average &                21.65 &              21.76 &                50.98 &               33.79 &               39.69 &             52.07 &              56.03 &                42.6 &                 50.52 &    44.55 &          13.31 \\
ar      &                43.56 &              18.03 &                51.12 &               34.34 &               38.93 &             53.02 &               55.0 &               39.26 &                 47.73 &    45.13 &           9.75 \\
bn      &                 1.25 &              10.53 &                34.53 &               20.92 &               22.61 &             49.11 &              50.58 &                2.22 &                  5.22 &    20.75 &           0.19 \\
cs      &                  9.8 &              21.76 &                50.23 &               33.63 &               38.13 &             49.81 &              53.78 &               45.08 &                 53.41 &    46.95 &          14.57 \\
da      &                 12.0 &              27.11 &                61.85 &               42.86 &               50.56 &             58.41 &              63.35 &               52.29 &                 62.73 &    54.32 &          18.57 \\
de      &                30.17 &              28.69 &                65.79 &               45.13 &                53.2 &             63.79 &              68.32 &               63.68 &                 72.44 &    67.06 &          19.62 \\
el      &                 4.08 &              19.64 &                51.76 &               35.51 &               41.45 &             49.07 &              55.02 &                45.0 &                 54.08 &    40.44 &          12.49 \\
es      &                48.75 &              24.24 &                55.36 &                37.6 &               43.55 &             54.41 &              58.22 &               54.02 &                 61.52 &    59.53 &          17.49 \\
fa      &                13.04 &               23.3 &                50.89 &               32.71 &                38.0 &             53.04 &              56.65 &               48.31 &                 56.69 &    51.36 &          13.02 \\
fi      &                 5.79 &              24.16 &                 58.1 &               38.04 &               44.02 &             57.68 &              63.25 &               42.95 &                 57.15 &    45.03 &          13.43 \\
fil     &                 7.29 &              16.92 &                45.98 &               29.26 &               34.87 &             43.71 &               47.9 &                7.27 &                  9.54 &    20.95 &           2.24 \\
fr      &                 55.4 &               28.4 &                62.51 &                42.2 &               50.36 &             62.24 &              66.63 &               60.79 &                 69.72 &    64.53 &           21.3 \\
he      &                 6.57 &               23.9 &                48.75 &               30.39 &               37.82 &             56.97 &              61.44 &               48.17 &                 57.93 &    51.01 &          10.99 \\
hi      &                 2.76 &               8.39 &                 29.9 &               15.01 &               18.52 &              34.8 &              36.21 &               17.12 &                  21.8 &    17.85 &           4.16 \\
hr      &                 8.53 &              29.31 &                61.41 &               41.96 &               48.72 &             57.51 &              62.01 &               48.82 &                 59.68 &    48.63 &          18.97 \\
hu      &                 9.95 &              23.31 &                62.83 &               43.35 &               50.98 &             56.93 &              63.29 &               49.74 &                 62.71 &    54.88 &           14.3 \\
id      &                18.52 &              28.91 &                 65.0 &               44.88 &               52.13 &             64.12 &               67.3 &                56.2 &                 66.29 &    61.03 &           21.5 \\
it      &                51.82 &              26.33 &                61.35 &               41.25 &               48.28 &             59.26 &              64.18 &               58.92 &                 67.19 &    63.49 &          16.49 \\
ja      &                58.65 &               27.1 &                54.15 &               33.83 &               40.38 &             58.29 &              61.64 &               59.14 &                 69.55 &    42.66 &          16.19 \\
ko      &                51.32 &              20.17 &                46.07 &               29.28 &               35.83 &             52.08 &              56.59 &               44.51 &                 53.44 &    47.58 &           9.74 \\
mi      &                 0.23 &               0.02 &                 0.34 &                0.13 &                0.27 &             23.99 &              27.68 &                0.51 &                  0.49 &     0.32 &           0.11 \\
nl      &                21.04 &              23.59 &                53.11 &               36.25 &               42.42 &             51.74 &               56.6 &               48.36 &                 56.37 &     52.8 &          16.93 \\
no      &                12.91 &              26.36 &                 58.1 &               39.46 &               46.32 &             57.35 &              61.93 &               49.98 &                 59.68 &    50.95 &          15.53 \\
pl      &                13.68 &              26.26 &                59.03 &               41.27 &               48.45 &             56.62 &              62.74 &               55.31 &                  65.2 &    58.03 &          17.98 \\
pt      &                38.08 &              25.18 &                57.08 &               38.98 &               46.62 &             55.65 &              60.97 &                55.2 &                 64.46 &    60.15 &          20.02 \\
quz     &                 2.69 &               0.81 &                 2.14 &                1.19 &                1.61 &             16.63 &               18.6 &                2.93 &                   3.5 &     3.03 &           0.75 \\
ro      &                17.35 &              25.19 &                64.59 &               42.82 &               51.38 &             59.85 &              65.96 &               56.32 &                 66.59 &    53.74 &          16.64 \\
ru      &                56.54 &              28.83 &                65.17 &               44.24 &               51.92 &             61.83 &              65.75 &               63.61 &                 72.36 &    68.18 &          20.17 \\
sv      &                12.83 &              25.38 &                59.29 &               40.71 &               47.63 &             55.29 &              60.36 &               51.57 &                 59.66 &    53.13 &          16.44 \\
sw      &                  2.5 &              12.52 &                39.02 &               22.75 &               27.59 &             41.98 &              44.78 &                 2.5 &                  3.18 &     11.1 &           0.41 \\
te      &                 3.86 &              13.18 &                26.83 &               14.54 &               17.24 &             37.71 &              39.99 &                0.25 &                  0.58 &     4.35 &           0.06 \\
th      &                13.38 &              20.43 &                52.29 &               33.21 &               38.72 &             53.87 &              56.49 &               44.44 &                 54.29 &    25.01 &           9.38 \\
tr      &                 7.51 &              22.98 &                55.25 &               37.65 &               44.37 &             54.27 &              57.83 &               46.41 &                 57.68 &    49.79 &          13.11 \\
uk      &                33.67 &              26.32 &                61.75 &               42.83 &                49.6 &             57.44 &              61.98 &               55.37 &                 65.18 &    54.87 &          17.24 \\
vi      &                 5.52 &               24.9 &                59.41 &               38.84 &               45.59 &             55.99 &              59.09 &               53.56 &                  64.2 &    52.67 &          17.25 \\
zh      &                54.74 &              26.44 &                56.97 &               37.59 &               44.05 &             51.98 &              54.96 &               55.34 &                 61.92 &    50.32 &          16.67 \\
\bottomrule
\end{tabular}
}
    \caption{XM3600 T2I R@1 results. Average is without English.}
    \label{tab:res:xm3600}
\end{table*}

\begin{table*}[]
    \centering
    \footnotesize

     \def\arraystretch{0.97}
     \resizebox{1\linewidth}{!}{
\begin{tabular}{lrrrrrrrrrrrr}
\toprule
            model &     en &     de &     es &     fr &     it &     jp &     ko &     pl &     ru &     tr &     zh &  average \\
\midrule
 AltCLIP XLMR-L L-14 & 64.4 & 36.0 & 58.6 & 60.1 & 57.8 & 53.3 & 56.7 & 17.7 & 53.9 & 10.7 & 58.6 &    46.34 \\
   M-CLIP mBERT B-32 & 44.5 & 40.9 & 41.4 & 42.0 & 41.5 & 33.4 & 35.5 & 41.4 & 35.4 & 39.0 & 40.7 &    39.12 \\
 M-CLIP XLMR-L B-16+ & 63.2 & 61.4 & 59.8 & 59.3 & 61.0 & 48.3 & 49.8 & 64.0 & 54.8 & 59.6 & 58.8 &    57.68 \\
  M-CLIP XLMR-L B-32 & 48.5 & 46.9 & 46.4 & 46.1 & 45.8 & 35.0 & 36.9 & 48.0 & 43.2 & 45.7 & 45.4 &    43.94 \\
  M-CLIP XLMR-L L-14 & 56.3 & 52.2 & 52.7 & 51.8 & 53.6 & 41.5 & 42.5 & 54.1 & 48.4 & 52.7 & 53.5 &    50.30 \\
    NLLB-SigLIP-base & 70.8 & 64.2 & 66.3 & 66.0 & 66.2 & 55.3 & 61.2 & 68.0 & 61.6 & 66.0 & 60.2 &    63.50 \\
   NLLB-SigLIP-large & 71.9 & 67.0 & 68.9 & 68.0 & 67.8 & 58.1 & 63.4 & 68.3 & 62.0 & 68.8 & 63.7 &    65.60 \\
  OpenCLIP XLMR B-32 & 63.2 & 54.5 & 54.6 & 55.7 & 55.7 & 47.1 & 43.8 & 55.5 & 50.3 & 48.2 & 50.8 &    51.62 \\
OpenCLIP XLMR-L H-14 & 73.5 & 64.8 & 65.9 & 64.7 & 64.9 & 64.3 & 56.4 & 68.7 & 62.4 & 62.7 & 61.9 &    63.67 \\
             mSigLIP & 68.0 & 60.8 & 62.7 & 59.9 & 58.1 & 32.9 & 49.6 & 59.6 & 56.9 & 55.9 & 50.4 &    54.68 \\
       ST mBERT B-32 & 43.3 & 32.3 & 32.9 & 32.0 & 29.8 & 21.4 & 18.5 & 28.8 & 25.5 & 25.3 & 31.0 &    27.75 \\
\bottomrule
\end{tabular}
}

    \caption{XTD T2I R@1 results. Average is without English.}
    \label{tab:res:xtd}
\end{table*}

\end{document}